\pdfoutput=1

\documentclass[11pt]{article}

\usepackage[preprint]{acl}

\usepackage{float}
\usepackage{relsize}
\usepackage{times}
\usepackage{latexsym}
\usepackage{microtype}
\usepackage{graphicx}
\usepackage{algorithm}
\usepackage{algorithmic}
\usepackage{subfigure}
\usepackage{booktabs} 
\usepackage{multirow}
\usepackage{caption}
\usepackage{bm}
\usepackage[labelsep=none]{caption}

\usepackage{hyperref}

\usepackage{amsmath}
\usepackage{amssymb}
\usepackage{mathtools}
\usepackage{amsthm}
\usepackage{array}
\usepackage{dsfont}

\theoremstyle{plain}

\theoremstyle{definition}

\theoremstyle{remark}

\usepackage[textsize=tiny]{todonotes}

\DeclareMathOperator*{\argmin}{\arg\!\min}

\usepackage[T1]{fontenc}

\usepackage[utf8]{inputenc}

\usepackage{microtype}

\usepackage{inconsolata}

\usepackage{graphicx}

\title{On Debiasing Text Embeddings Through Context Injection}

\author{Thomas Uriot \\
  NATO Communications and Information Agency (NCIA) \\
  \texttt{thomas.uriot@gmail.com}}

\begin{document}
\maketitle
\begin{abstract}
Current advances in Natural Language Processing (NLP) have made it increasingly feasible to build applications leveraging textual data. Generally, the core of these applications rely on having a good semantic representation of text into vectors, via embedding models. However, it has been shown that these embeddings capture and perpetuate biases already present in text. While a few techniques have been proposed to debias embeddings, they do not take advantage of the recent advances in context understanding of modern embedding models. In this paper, we fill this gap by conducting a review of 19 embedding models by quantifying their biases and how well they respond to context injection as a mean of debiasing. We show that higher performing models are more prone to capturing biases, but are also better at incorporating context. Surprisingly, we find that while models can easily embed affirmative semantics, they fail at embedding neutral semantics. Finally, in a retrieval task, we show that biases in embeddings can lead to non-desirable outcomes. We use our new-found insights to design a simple algorithm for top $k$ retrieval, where $k$ is dynamically selected. We show that our algorithm is able to retrieve all relevant gendered and neutral chunks. 

\end{abstract}

\section{Introduction}
\label{introduction}

Recently, the use of Artificial Intelligence (AI) in text-based applications has increased dramatically. This widespread adoption has mainly been due to the rise of Large Language Models (LLMs). While LLMs have shown great potential in tackling previously unsolvable tasks, a large portion of text-based applications is solely based on embeddings. Such tasks include document classification \cite{stein2019analysis}, topic representation and clustering \cite{grootendorst2022bertopic}, and information retrieval \cite{thakur2021beir}. Since embedding models form the basis of so many NLP applications, it is crucial to

\begin{figure}[H]
    \hspace{-1.9em}
    \begin{minipage}[b]{0.57\textwidth}
        \includegraphics[width=\textwidth]{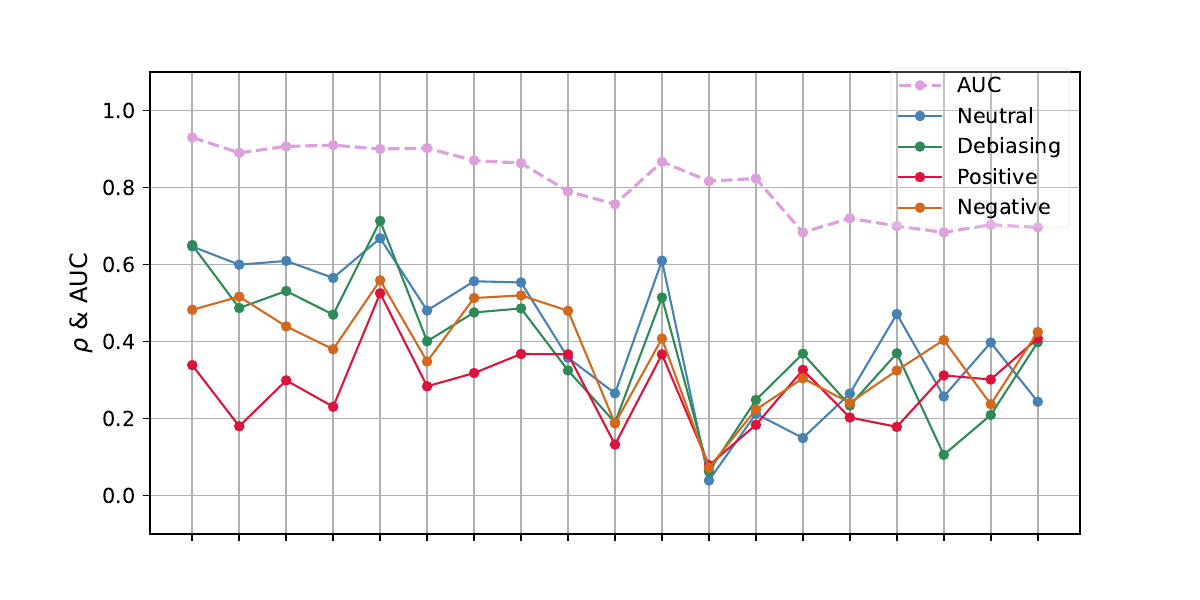}
    \end{minipage}
    \hspace*{-1.3em}
    \begin{minipage}[b]{0.505\textwidth}
        \includegraphics[width=\textwidth]{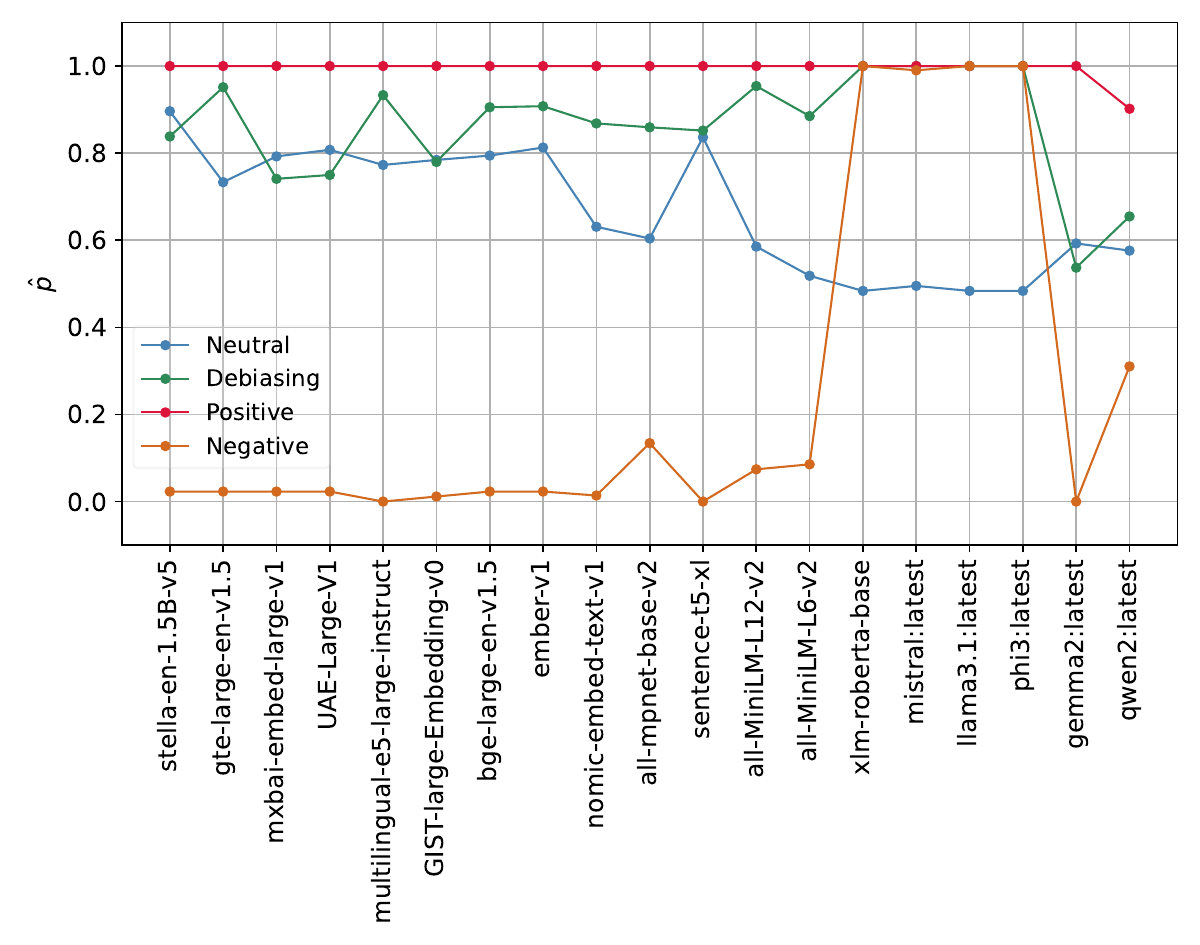} 
    \end{minipage}
   \caption{:\hspace{0.1cm}\textbf{Top -} Dashed line shows the averaged AUC scores (see Equation \ref{eq:auc}) across the learned concepts. A higher AUC means a better concept representation: we can see that stronger models (higher ranked on the MTEB) have learned a better concept direction ($\rho=0.76$). The solid lines shows the average correlation (a proxy for bias strength) across the three concepts (gender, age, wealth), between human-labeled attributes and their projections onto the concept directions. The blue line shows that stronger models are more prone to containing biases (\textit{neutral}, $\rho=0.79$).
   \textbf{Bottom -} The solid lines show the sample proportion $\hat{p}$ of the binomial test. The blue line can be interpreted as a measure of bias present in the embeddings. Models higher ranked on the MTEB are more prone to containing biases ($\rho = 0.77$). In green, values closer to $\hat{p}=0.5$ means that debiasing has been most effective. The red and yellow lines show the ability of an embedding to correctly capture ``affirmative'' semantics. Several models (yellow line) fails to capture negative semantics.
   }
    \label{fig:method1_final_results}
\end{figure}

objectively evaluate their performance. To this end, the Massive Text Embedding Benchmark (MTEB) \cite{muennighoff2022mteb}, hosted on HuggingFace\footnote{\url{https://huggingface.co/spaces/mteb/leaderboard}}, ranks the models across 7 tasks and 56 datasets. While the MTEB is a great resource to find a model that corresponds to our application needs and hardware requirements, it ignores a key aspect: these models, like any machine learning system trained on real-world data, are prone to learn the biases present in their training data. 

This is particularly problematic for textual data as it may contain societal biases and stereotypes. Blindly applying models would risk to perpetuate and exacerbate these biases, as well as lead to erroneous and incomplete results, as shown in Section \ref{sec:toy_rag}. This issue is not new as biases were already present in the earliest implementations of learned word representations Word2Vec \cite{mikolov2013efficient} and GloVe \cite{pennington2014glove}. Indeed, it was shown that Word2Vec not only captures natural concepts such a gender \cite{mikolov2013linguistic}, but also biases related to gender such as occupational bias \cite{bolukbasi2016man}. To identify gender-related biases, the authors \cite{bolukbasi2016man} first find a subspace representing the gender direction (vector), which is then used to project embedded words of interest (e.g., nurse, electrician) onto. However, this method rely on human-labels and requires a carefully curated set of words in order to represent a given concept (e.g., gender, age, race). On the other hand, in \cite{caliskan2017semantics}, the authors introduced a largely unsupervised algorithm to discover biases automatically from an unlabeled data representation. This method, called Word Embedding Association Test (WEAT). It only requires two groups of target words (e.g., flowers vs insects) and two groups of attribute words (e.g., pleasant vs unpleasant). One can then perform a Wilcoxon signed-rank test or a binomial test in order to assess whether or not there is a significant difference between the two groups (see Section \ref{sec:method_identify_bias}). While these methods are, in principle, applicable to arbitrary embedding models, these works only looked at basic, non-contextual, word-level embeddings (Word2Vec, GloVe, fastText \cite{bojanowski2017enriching}).

Effectively identifying biases allows us to put in place potential safeguards in our applications and steer outputs to more acceptable and fair outcomes. Several methods to debias directly at the embeddings level have been recently proposed. In \cite{bolukbasi2016man}, the authors proposed a debiasing method motivated by geometry whereby neutral words should be equidistant to non-neutral words (e.g., gendered words). However, this method has been shown to still perpetuate biases \cite{swinger2019biases}, and is not easily applicable to debiasing several concepts at once. In addition, traditional debiasing techniques \cite{bolukbasi2016man, zhang2018mitigating, manzini2019black}, whether it be post-hoc debiasing or during the training of the embeddings, have been shown to ``hide'' the bias rather than removing it \cite{gonen2019lipstick, prost2019debiasing}. More recently, several studies \cite{zhao2019gender, kaneko2021debiasing} have looked into bias in more modern, context-aware embedding models such as BERT \cite{devlin2018bert}, Roberta \cite{liu2019roberta} and ELMo and have also found the presence of biases in these embeddings. Similarly, deabiasing techniques for contextualized embeddings have been proposed. In \cite{zhao2019gender}, the authors propose to swap gender (e.g., she-he) in sentences, and take the average of the two distinct embeddings as the final representation. However, this is not very practical as one would have to first identify all gender-bearing words and replace them by an appropriate counterpart while preserving meaning. Furthermore, it is unclear whether the resulting embedding would still make sense. Instead, in \cite{kaneko2021debiasing}, the authors propose a fine-tuning approach where the loss function forces the embeddings to be orthogonal to a learned concept direction (e.g., gender). This technique still requires to accurately define the concept direction, and is confronted to balancing debiasing vs preserving information during the fine-tuning process. 

In the last couple of years, there has been an explosion of contextualized embedding models, fueled by the introduction of the MTEB. These modern models are trained to capture the semantic meaning of text at sentence or even paragraph-level. Several top-ranked models in the MTEB are now based on pre-trained LLMs \cite{behnamghader2024llm2vec, li2023towards, lee2024nv, meng2024sfrembedding}. With the introduction of such powerful embedding models, it is worth investigating whether adding context could work as a way to debiase embeddings. 

In this work, we aim to fill this knowledge gap by providing a systematic analysis of 19 embbedding models chosen from the MTEB leaderboard (see Table \ref{table:models}), ranging from LLMs by taking the last \texttt{<EOS>} token embedding, to SOTA specialized embeddings models, as well as less performing yet popular models. To do so, we first quantify the models' biases by modifying two existing techniques: (i) geometrical projection \cite{bolukbasi2016man}, and, (ii) WEATS \cite{caliskan2017semantics}. With p-values obtained from our analysis, we can  also assess the effect of using context injection to debias embeddings. In particular, we seek to test two hypotheses:

\begin{enumerate}
    \item Better performing models are more prone to capturing biases related to concepts such as gender, age, and wealth.
    \item Better performing models are more responsive to context added for debiasing embeddings.
\end{enumerate}

To the best of our knowledge, our main contributions are novel and can be summarized as follows:


\begin{itemize}
    \item We conduct the first systematic review of 19 embedding models, quantifying their biases using two techniques and relate bias strength to the models' MTEB rankings. We also introduce a way to measure if a pre-defined concept (e.g., gender, age) is captured by a single vector.
    \item We introduce a novel and simple debiasing technique using context injection. We evaluate each model's response under various contexts to assess the viability of the method.
    \item Finally, we showcase how biases in embeddings can impact a simple retrieval system and present an algorithm using context injection to mitigate the issue. Our algorithm is based on top-$k$ retrieval with $k$ chosen dynamically.
\end{itemize}

\section{Methodology}
\label{sec:methdology}

As a piece of notation, we will denote raw text as strings in double quotes, and as a shorthand, we will refer to the embeddings matrix as $X\in \mathbb{R}^{n \times p}$, where $x_{i} \in \mathbb{R}^{p}$ corresponds to the $i^{th}$ row of $X$ and represents a text embedding.

 \subsection{Identifying Biases}
\label{sec:method_identify_bias}
As mentioned, we apply two prominent techniques from the literature in order to identify biases: (i) a geometry-based approach \cite{bolukbasi2016man}, and, (ii) an approach directly based on distances between pair of words \cite{caliskan2017semantics}. These two approaches both have their pros-and-cons, which will be discussed in the next section.
\subsubsection{Geometrical bias identification}
\label{sec:method_geometry_identify_bias}

This method aims to first identify a subspace related to a predefined concept (e.g., gender, age, race) and to then project biased (or potentially biased) terms onto this direction (e.g., occupations such as nurse or electrician). The authors denote this concept subspace as \( B = \{b_1, \ldots, b_k\} \subset \mathbb{R}^p \), consisting of \( k \) orthogonal unit vectors.

To identify the concept subspace, we require a set of opposing word pairs related to the concept we are interested in. For instance, for the concept of gender, such pairs could be \textit{she-he}, \textit{mother-father}, or \textit{aunt-uncle} whose definitions differ mainly in gender. Formally, we define these two sets of $n$ words as $W_{1}$ and $W_{2}$, and their corresponding embeddings as $X_{1} \in \mathbb{R}^{n \times p}$ and $X_{2} \in \mathbb{R}^{n \times p}$. In addition, we require a set of attributes whose embeddings are denoted by $A \in \mathbb{R}^{m \times p}$, and their corresponding human-annotated labels $y \in \mathbb{R}^{m}$, $y \in [0, 1]$, representing the bias strength. For instance, in the case of gender, a typical set of words could be occupations, where terms with a more female connotation (e.g., nurse, homemaker) would be closer to 0, and closer to 1 for male (e.g., miner, electrician). The data used in the paper for this method can be found in Appendix \ref{appendix:data_geometry}.

Further, we denote the element-wise difference in embeddings as $\Delta X = X_2 - X_1$. Intuitively, the embeddings in $\Delta X$ represent the concept difference (e.g., gender) across several data pairs. However, this is a noisy process and each row of the $\Delta X$ matrix will differ. The goal is then to extract a meaningful subspace $B$, from $\Delta X$ which will represent our concept. In this method, we take the singular value decomposition (SVD) of $\Delta X$ and to keep the top $k$ eigenvectors as the concept subspace. To simplify the subspace to a single direction (vector), the authors assume that the first eigenvector $b_1$ (with the highest associated eigenvalue) will best represent the concept. Thus, we have that the concept direction, denoted by $g$, is simply given by $g=b_1$. 
Note that in practice we often have $n \ll p$ due to the high-dimensional nature of embeddings and to the difficulty in constructing the sets $W_{1}$ and $W_{2}$. While this may seem like a major issue, the intrinsic dimensionality of $\Delta X$ can be much smaller than $p$. In our paper, we extend the method and we show whether or not the direction $g$ has truly captured the concept we are interested in. We do so by projecting $X_2$ and $X_1$ onto $g$ and compute the Area Under the Curve (AUC) in a binary classification setting. While the authors in \cite{bolukbasi2016man} use $b_1$ as their concept direction by default, we have found (see Appendix \ref{appendix:auc}) that it is not always the best choice (i.e., not the direction with the highest AUC). Thus, in this paper, we want $g$ such that

\begin{equation}
\label{eq:auc}
g = \arg\max_{b \in B} \text{AUC}(b),
\end{equation}

where \(\text{AUC}(b)\) denotes the AUC value achieved by a linear binary classifier whose goal is to classify the projections of the concept-defining pairs such as \textit{she-he}, \textit{mother-father}. Empirically, we find that when $\text{AUC}(g)<0.8$, the direction $g$ does not capture the concept well. Thus, any subsequent bias quantification becomes unreliable, as the concept was not learned in the first place (see top of Figure \ref{fig:method1_final_results}). Once the concept direction $g$ has been identified, we can now quantify the bias. This is done by taking the product between the attribute matrix $A$ and our concept $g$: $a_g = A g$, where $a_g \in \mathbb{R}^n$. Finally, the bias can be quantified as 

\begin{equation}
\label{eq:bias_corr}
    \rho=\text{corr}(y, a_g),
\end{equation}

where $y$ is the human-annotated labels representing bias strength (see Table \ref{tab:bias_labels} in Appendix \ref{appendix:data_geometry}). While $\rho$ represents the correlation between our output and the perceived bias of the annotators, how can we be sure that it did not arise by chance? This is especially important as $y$ contains only a few dozen values (see Table \ref{tab:bias_labels}). For statistical significance, we compute $\text{corr}(y, a_i)$, for $i=1,\ldots,N$, where $a_i \sim N(0,I)$ represent random projections. The p-value is defined as $p = P(T \geq |\rho|)$, and can be estimated using random samples as

\begin{equation}
    \label{eq:p_value_1}
    \hat{p} = \frac{1}{N}\sum_{i}^{N} \mathds{1}_{\text{corr}(y, a_i)\geq |\rho|}.
\end{equation}


The principal strength of this method is that we explicitly compute the concept direction. Ideally, this direction does not contain any semantic meaning other than our concept. This is very useful for downstream bias detection, debiasing, as well as quantifying the contribution of $g$ to the similarities between any pair of words \cite{bolukbasi2016man}. However, it can be difficult to construct $W_{1}$ and $W_{2}$ for more complex concepts, and requires human-annotated labels. Finally, the assumption that a concept can be reduced to a single eigenvector of $\Delta X$ and thus projected onto a single dimension may be too restrictive. Indeed, in our experiments, we have found that 2 dimensions (eigenvectors) often significantly outperforms using a single dimension.

\subsubsection{WEATs}
\label{sec:method_weat_identify_bias}

In contrast, WEAT is simpler to implement as it does not require $W_{1}$ and $W_{2}$. Instead, it requires two pairs of sets. The first pair is referred to as the \textit{targets}, and the second pair as the \textit{attributes}. In addition, WEAT is computed in terms of the cosine similarity between the original embeddings. In that sense, it is a better representation of what can be expected in real-world applications (e.g., RAG, clustering).

Let us denote the target sets as $T_1$ and $T_2$, and similarly, $A_1$ and $A_2$ for the attribute sets. For example, for gender and occupations, we might have $T_1 =$ \{she, aunt, \ldots, feminine\}, $T_2 =$ \{father, uncle, \ldots, beard\} and $A_1 =$ \{nurse, homemaker, \ldots, social worker\}, $A_2 =$ \{miner, electrician, \ldots, builder\}. Note that instead of labeling the attributes between 0 and 1 (as was the case for the previous method), we now only have to put the attributes into two classes. In addition, we do not require $T_1$ and $T_2$ to be precise counterparts, allowing for greater flexibility in the topics and concepts we want to test. The corresponding embeddings are denoted by $X_1$, $X_2$ for the target sets and $Y_1$ and $Y_2$ for the attribute sets. In WEAT, we want to test the null hypothesis $H_0$ that there is no difference between the two sets of target words in terms of their relative similarity to the two sets of attribute words \cite{caliskan2017semantics}. To test this hypothesis, we make the assumption and construct the sets in such a way that target terms in $T_1$ should be more related to attributes in $A_1$, similarly for $T_2$ and $A_2$. Then, we want to compute the ordered sequences:

\begin{equation}
     \fontsize{10}{12}
    \begin{aligned}
        S_{T_j, A_j} &= \left( \frac{1}{|T_j|} \sum_{t \in T_j} \cos(a_i, t) \;\middle|\; a_i \in A_j,\; j \in \{1,2\} \right), \\
        S_{T_2, A_1} &= \left( \frac{1}{|T_2|} \sum_{t \in T_2} \cos(a_i, t) \;\middle|\; a_i \in A_1 \right), \\
        S_{T_1, A_2} &= \left( \frac{1}{|T_1|} \sum_{t \in T_1} \cos(a_i, t) \;\middle|\; a_i \in A_2 \right).
    \end{aligned}
\end{equation}

\newpage

In words, $S_{T_j, A_j}$, $j = 1, 2$, are the two sequences where we take the similarity between the related target and attribute sets ($T_1$, $A_1$), and ($T_2$, $A_2$). On the other hand $S_{T_1, A_2}$ and $S_{T_2, A_1}$ can be seen as the opposite where we take the similarity of contrasting (or non-related) terms. In the remainder of this section, we will denote $s^{l}_{i, j}$ to be the $l^{th}$ element of sequence $S_{T_i, A_j}$. We are now equipped to devise several hypothesis tests based on the binomial test statistic. We set-up a different test for each of the scenarios represented in Table \ref{table:scenarios_gender} (see also Table \ref{table:scenarios_age} and Table \ref{table:scenarios_wealth}). These tests are described in Table \ref{table:statistical_testing}, where the number of successes $k_1$ and $k_2$ in are given by:

\begin{equation}
\label{eq:successes_k1}
k_1 = \sum_{i=1}^{|A_1|} \mathds{1}_{s_{1,1}^{(i)} - s_{2,1}^{(i)} > 0} + \sum_{i=1}^{|A_2|} \mathds{1}_{s_{2,2}^{(i)} - s_{1,2}^{(i)} > 0},
\end{equation}

\begin{equation}
\label{eq:successes_k2}
k_2 = \sum_{i=1}^{|A_1|} \mathds{1}_{s_{1,1}^{(i)} - s_{2,1}^{(i)} > 0} + \sum_{i=1}^{|A_2|} \mathds{1}_{s_{1,2}^{(i)} - s_{2,2}^{(i)} > 0}.
\end{equation}

\begin{table}[h!]
\centering
\begin{tabular}{llcc}
\specialrule{1.3pt}{0pt}{0pt} 
\textbf{Scenarios} & {\textbf{$H_0$}}  & {\textbf{$H_1$}} & \textbf{k}\\
\toprule
\vspace{0.1cm}
\multirow{1}{*}{\textbf{Neutral}}&  $p=\frac{1}{2}$ & greater & $k_{1}$ \\
\vspace{0.1cm}
\multirow{2}{*}{\textbf{Debiasing}} &  $p=\frac{1}{2}$ & two-sided & $k_{1}$ \\
\vspace{0.1cm}
&  $p=\frac{|A_1|}{|A_1|+|A_2|}$ & greater & $k_{2}$ \\
\vspace{0.1cm}
\multirow{1}{*}{\textbf{Positive}} &  $p=\frac{|A_1|}{|A_1|+|A_2|}$ & greater & $k_{2}$ \\
\vspace{0.1cm}
\multirow{1}{*}{\textbf{Negative}} &  $p=\frac{|A_1|}{|A_1|+|A_2|}$ & less & $k_{2}$ \\
\specialrule{1.3pt}{0pt}{0pt} 
\end{tabular}
\caption{:\hspace{0.1cm}Null hypothesis $H_0$, the alternative $H_1$, and the number of successes $k$, for each of the binomial tests, depending on the scenario.}
\label{table:statistical_testing}
\end{table}

Simply put, for the \textit{neutral} scenario, we are testing whether or not the similarities are closer between our (assumed to be) related two groups, compared to the other two. If $A_1$ is no closer to $T_1$, and $A_2$ to $T_2$, then we would roughly expect to observe $\hat{p}=\frac{1}{2}$. If we observe $\hat{p}$ to be significantly larger than $\frac{1}{2}$, we reject the null hypothesis of no bias in favor of bias. For \textit{debiasing}, we perform two tests: (i) firstly, we test whether $\hat{p}$ departs from $\frac{1}{2}$. We are equally interested in whether the context has no effect when compared to \textit{neutral} ($H_1: p > \frac{1}{2}$) or whether it successfully nudges $\hat{p}$ lower ($H_1: p < \frac{1}{2}$), hence a two-sided test. (ii) Secondly, if we rejected $H_0$ in the first test due to the lower-tail ($H_1: p < \frac{1}{2}$), we then want to test if this was due to a real ``debiasing'' effect or because all attributes from $A_1$ and $A_2$ are now both closer to $T_1$. For example, for the gender concept, this would mean that all occupations (e.g., nurse, electrician) would be closer to the female terms, similarly to the \textit{positive} case. In a way, instead of debiasing, the model would ``over-compensate'' in favor of the discriminated/biased against group (female in this case). Therefore, when displaying results for \textit{debiasing} in Figure \ref{fig:method1_final_results} (bottom) and Table \ref{table:method_2_results}, we display $k = \text{max}(k_1, k_2)$. A large $k_1$ means debiasing has no effect while a large $k_2$ means it had an ``over-compensating'' effect. For debiasing to have had its wanted effect, we would like to observe a lower $k$ value than for \textit{neutral}. Finally, for the positive and negative scenarios, we are testing whether the statistic becomes skewed towards the group (e.g., male, female) mentioned in the context.

\subsection{Debiasing through context injection}
\label{sec:method_debias}

Modern, SOTA contextualized embedding models are often advertised as truly capturing the semantic meaning of text at sentence or paragraph-level. However, as a cautionary tale, it has been shown that SOTA embedding models such as \textit{UAE-Large-v1} \cite{li2023angle}, \textit{Sentence-T5-x} \cite{ni2021sentence}, and even models based on pre-trained LLMs \cite{wang2023improving} fail to truly dissociate the semantic meaning from the lexical similarities of text \cite{hsieh2024sugarcrepe, harsha2024sugarcrepe++}. Nonetheless, we set-out to test whether or not SOTA models are able to embed the semantics contained in text as a mean to debasing. The idea is that if these models were to truly embed semantics, the sentence \textit{``This person is a nurse.''} may indeed be biased towards the female gender, while the sentence \textit{``This person is a nurse. It could be a man or a woman.''} should not carry such bias whilst conveying the same meaning. Thus, we craft simple prompts, that we will be referring as ``contexts'', to be applied to each of the concepts we want to test for bias. The prompts for the \textit{gender} concept are shown in Table \ref{table:scenarios_gender}, and the prompts for the \textit{age}, and \textit{wealth} concepts can be found in Appendix \ref{appendix:prompts} in Table \ref{table:scenarios_age}. The \{attribute\} placeholder will depend on the concept we are interested in. For instance, \textit{occupations} for gender and \textit{character traits} for age (see Table \ref{tab:bias_labels} in Appendix \ref{appendix:data_geometry}). In addition to adding a debiasing context, we also add a \textit{positive}, and a \textit{negative} context. These contexts will be used to assess whether it is easier for embedding models to accurately capture the semantics of an affirmative context (positive or negative) as opposed to a neutral one.

\setlength{\textfloatsep}{20pt} 

\begin{table}[h!]
\centering
\begin{tabular}{p{1.5cm}p{5.3cm}}
\specialrule{1.3pt}{0pt}{0pt} 
\textbf{Scenarios} & \multicolumn{1}{c}{\textbf{Text to embed}} \\
\specialrule{0.5pt}{0pt}{0pt} 
\multirow{3}{=}{\textbf{Neutral}} & I want to find information about a person. This person's profession is \{attribute\}. \\
\specialrule{0.5pt}{0pt}{0pt} 
\multirow{4}{=}{\textbf{Debiasing}} & I want to find information about a person. This person's profession is \{attribute\}. This person's gender is unknown.  \\
\specialrule{0.5pt}{0pt}{0pt} 
\multirow{4}{=}{\textbf{Positive}} & I want to find information about a person. This person's profession is \{attribute\}. This person is a woman.  \\
\specialrule{0.5pt}{0pt}{0pt} 
\multirow{4}{=}{\textbf{Negative}} & I want to find information about a person. This person's profession is \{attribute\}. This person is a man.  
\\
\specialrule{1.3pt}{0pt}{0pt} 
\end{tabular}
\caption{:\hspace{0.1cm}Example of the four scenarios, for the gender concept. Note that the use of ``positive'' and ``negative'' is arbitrary and not to be taken literally. }
\label{table:scenarios_gender}
\end{table}

\section{Experiments}
\label{sec:experiments}

\subsection{Embedding Models}
\label{sec:models}

We experiment with 19 models, out of which 14 are specialized embedding models, and 5 are LLMs. We can group these models into 4 categories: (i) a top-3 model with 1.5B parameters, (ii) SOTA models (rank 21 to 54), (iii) less performing, yet still popular models (rank 116 to 134), and (iv) LLMs.

\begin{table}[h!]
\centering
\begin{tabular}{|c|c|}
\textbf{Embedding models} & \textbf{score (rank)} \\
\hline
stella-en-1.5B-v5 & 71.19 \hspace{0.1cm}(3) \\
gte-large-en-v1.5 & 65.39\hspace{0.1cm} (21) \\
mxbai-embed-large-v1 & 64.68 \hspace{0.1cm}(25)\\
UAE-Large-V1 & 64.64 \hspace{0.1cm}(26) \\
multilingual-e5-large-instruct & 64.41\hspace{0.1cm} (30) \\
GIST-large-Embedding-v0 & 64.34 \hspace{0.1cm}(32) \\
bge-large-en-v1.5 & 64.23\hspace{0.1cm} (33)  \\
ember-v1 & 63.34\hspace{0.1cm} (46)  \\
nomic-embed-text-v1 & 62.39\hspace{0.1cm} (54) \\
sentence-t5-xl & 57.87\hspace{0.1cm} (116)  \\
all-mpnet-base-v2 & 57.77\hspace{0.1cm} (117) \\
all-MiniLM-L12-v2 & 56.46\hspace{0.1cm} (127) \\
all-MiniLM-L6-v2 & 56.09 \hspace{0.1cm}(134)  \\
xlm-roberta-base & n/a  \\
mistral & n/a \\
phi3 & n/a \\
qwen2 & n/a \\
gemma2 & n/a \\
llama3.1 & n/a \\
\end{tabular}
\caption{:\hspace{0.1cm}Embedding models with their associated MTEB score and rank (as of the $20^{th}$ of September 2024). Models that do not have a MTEB score are marked with n/a.}
\label{table:models}
\end{table}

\subsection{Results}
\label{sec:results}

In our experiments, we apply the two bias quantification algorithms described in Section \ref{sec:methdology}, for each concept, and the 4 different contexts. For the two methods, the bias strength will be measured by a p-value. By using a p-value, we are able to assess whether or not bias is present in the first place, and if so, whether debiasing was successful.

\subsubsection{Geometrical bias identification}

From Figure \ref{fig:method1_final_results}, we can see that the human-labeled attributes and the projection of those attributes onto the concept direction are highly correlated (more detailed results in Table \ref{table:method_1_results} in Appendix \ref{sec:appendix_method1}). This is especially true for the better performing models, confirming our initial hypothesis that higher-ranked models are more prone to capturing biases. We can also see that the AUC, as described in \ref{eq:auc}, is higher for more performing models. These observations go hand-in-hand as a likely explanation may be that the higher the model is ranked, the more intricate relationships and concepts it is able to capture (e.g., the interaction between gender and occupations). On the other hand, we can see that results for \textit{debiasing} context remains strongly tied to \textit{neutral} context, suggesting that it has little effect on bias. While this bias detection method provides us with valuable insights, it relies on the concept subspace to be reduced to a single direction, while still capturing the concept well (i.e., high AUC). Finally, we are also relying on noisy continuous human-labeled data. These aforementioned limitations can be alleviated by using WEAT.

\begin{table*}[t]
\centering
\resizebox{0.84\textwidth}{!}{
\begin{tabular}{lllllllllll}
\toprule
 \multicolumn{3}{c}{\shortstack{\textbf{Concept = gender}, \\ \textbf{Attributes = occupations}}} & \multicolumn{3}{c}{\shortstack{\textbf{Concept = age}, \\ \textbf{Attributes = character traits}}} &  \multicolumn{3}{c}{\shortstack{\textbf{Concept = wealth}, \\ \textbf{Attributes = ethnicities}}}\\
\cmidrule(lr){1-3} \cmidrule(lr){4-6} \cmidrule(lr){7-9} 
 \textbf{neutral} & \textbf{debiasing} & \textbf{negative} & \textbf{neutral} & \textbf{debiasing} & \textbf{negative} &  \textbf{neutral} & \textbf{debiasing} & \textbf{negative} \\
\midrule

$0.96^{***}$ & $0.69^{**}$ & $0.0^{***}$ & $0.96^{***}$ & $0.97^{***}$ & $0.07^{***}$ &$0.76^{***}$& $1.0^{***}$&$0.0^{***}$\\
$0.94^{***}$ & $0.85^{***}$ & $0.0^{***}$ & $0.79^{***}$ & $0.58$ & $0.07^{***}$ & $0.67^{**}$&$1.0^{***}$&$0.0^{***}$\\
$0.94^{***}$ & $0.73^{***}$ &$0.0^{***}$ & $0.76^{***}$ & $0.79^{***}$ & $0.07^{***}$ &$0.64^{*}$&$0.73^{***}$&$0.0^{***}$\\
$0.96^{***}$ & $0.85^{***}$ & $0.0^{***}$ & $0.79^{***}$ & $0.76^{***}$ & $0.07^{***}$ &$0.71^{**}$&$0.71^{**}$&$0.0^{***}$\\
$0.94^{***}$ & $0.94^{***}$ & $0.0^{***}$ & $0.83^{***}$ & $0.79^{***}$ & $0.0^{***}$ &$0.59$&$1.0^{***}$&$0.0^{***}$\\
$0.94^{***}$ & $0.88^{***}$ & $0.0^{***}$ & $0.69^{**}$ & $0.83^{***}$ & $0.03^{***}$ &$0.59$&$0.71^{**}$&$0.0^{***}$\\
$0.96^{***}$ & $0.85^{***}$ & $0.0^{***}$ & $0.72^{**}$ & $0.69^{**}$ & $0.07^{***}$ &$0.74^{***}$&$1.0^{***}$&$0.0^{***}$ \\
$0.98^{***}$ & $0.89^{***}$ & $0.0^{***}$ & $0.66^{*}$ & $0.72^{**}$ & $0.07^{***}$ &$0.74^{***}$&$1.0^{***}$&$0.0^{***}$\\
$0.70^{***}$ & $0.71^{***}$ & $0.04^{***}$ & $0.69^{**}$ & $0.65^{*}$ & $0.0^{***}$ &$0.53$&$1.0^{***}$&$0.0^{***}$\\
$0.42$ & $1.0^{***}$ & $0.33^{**}$ & $0.86^{***}$ & $0.69^{**}$ & $0.07^{***}$ &$0.71^{**}$&$0.65$&$0.0^{***}$\\
$1.0^{***}$ & $0.79^{***}$ & $0.0^{***}$& $0.55$ & $0.86^{***}$ & $0.0^{***}$ & $0.65^{*}$&$0.97^{***}$&$0.0^{***}$\\
$0.65^{**}$ & $1.0^{***}$ & $0.19^{***}$ & $0.51$ & $0.55$ & $0.03^{***}$ &$0.56$&$1.0^{***}$&$0.0^{***}$\\
$0.48$ & $1.0^{***}$ & $0.19^{***}$ & $0.51$ & $0.52$ & $0.07^{***}$ &$0.56$&$1.0^{***}$&$0.0^{***}$\\
$0.38$ & $1.0^{***}$ & $1.0$ & $0.55$ & $0.52$ & $1.0$ &$0.56$&$1.0^{***}$&$1.0$\\

$0.38$ & $1.0^{***}$ & $1.0$ & $0.51$ & $0.55$ & $1.0$ &$0.56$&$1.0^{***}$&$0.97$\\
$0.38$ & $1.0^{***}$ & $1.0$& $0.51$ & $0.52$ & $1.0$ & $0.56$&$1.0^{***}$&$1.0$\\
$0.38$ & $1.0^{***}$ & $1.0$ & $0.51$ & $0.52$ & $1.0$ &$0.56$&$1.0^{***}$&$1.0$\\
$0.85^{***}$ & $0.69^{**}$ & $0.0^{***}$ & $0.48$ & $0.48$ & $0.0^{***}$ &$0.44$&$0.44$&$0.0^{***}$\\
$0.67^{**}$ & $0.83^{**}$ & $0.0^{***}$ & $0.62$ & $0.62$ & $0.93$ &$0.44$&$0.44$&$0.0^{***}$\\

\bottomrule
\end{tabular}
}
\caption{:\hspace{0.1cm}Sample proportion $\widehat{p}$ for the associated Binomial tests described in Table \ref{table:statistical_testing}, computed for three different contexts (neutral, debiasing, negative), across 3 pairs of concepts and attributes. The value for positive is always 1.0 (see bottom graph of Figure \ref{fig:method1_final_results}). The p-value is given by the asterisk notation. For a confidence level of $1 - \alpha$, statistical significance is denoted as follows: $^{*}, ^{**}, ^{***}$, for $\alpha$ = $10\%$, $5\%$, and $1\%$, respectively.}
\label{table:method_2_results}
\end{table*}

\subsubsection{WEATs}

Results for this method are displayed in Table \ref{table:method_2_results}. We can see that, similarly to the previous method, higher-ranked models are more prone to biases. We can also see that these models better capture the semantics of the added context, in particular for the \textit{negative} context. Indeed, 4 models completely fail at embedding \textit{negative} context, and in fact wrongly embed that context as \textit{positive}. Similarly, a surprising and important result is that \textit{debiasing} context often lead to having the opposite of the expected effect. This can be seen with $\hat{p}$ in Table \ref{table:method_2_results} and Figure \ref{fig:method1_final_results} where the results for the \textit{debiasing} context are similar to those of the \textit{positive} context ($\hat{p}$ close to 1). We dub this finding as the ``over-compensating effect" whereby the model actually steer the embeddings towards the discriminated against group. This signifies that, unfortunately, models do not accurately embed the \textit{debiasing} semantics as none of them have $\hat{p}$ close to $0.5$, the value for which debiasing would have been most effective. Finally, while all models (except \textit{qwen2}) score a perfect 1.0 for the \textit{positive} context, we see a decline in performance for the \textit{negative} context, especially towards the tail end of the model list.

\subsubsection{Common findings}

In this section, we look at patterns that have persisted across the two methods. First, we can see that the results for bias detection (\textit{neutral}) for both methods are highly correlated: $\rho=0.82$. On a more individual level, we can identify 9 models that exhibit high bias: the 8 higher-ranked models (from \textit{stella-en-1.5B-v5} to \textit{ember-v1}), and \textit{sentence-t5-xl}. Only 3 models seem to take into account the \textit{debiasing} semantics, principally for the \textit{gender} concept: \textit{mxbai-embed-large-v1}, \textit{UAE-Large-V1}, and \textit{gemma2}. Finally, for both methods, we see that the AUC is highly correlated to bias, which is in turn highly correlated to the MTEB ranking. This also stands for models in the LLMs group as they exhibit low AUC and low bias. This does not necessarily mean that these models produced quality unbiased embeddings, but instead that they do not capture subtle relationships in the data.

\section{Simple Retrieval Task}
\label{sec:toy_rag}

Inspired by our findings, we devise a toy retrieval task, akin to the first step used in a RAG application. We rank chunks according to their cosine similarities with respect to various queries (\textit{neutral}, \textit{debiasing}, \textit{positive}, \textit{negative} queries), shown in Appendix \ref{appendix:toy_retrieval}. We show that biases present in the embeddings lead to a skewed retrieval. We introduce a simple algorithm to ensure that our retrieval system is less biased and captures most of the relevant information. Figure \ref{fig:similiraty_matrix} below displays the similarity matrix between the queries and chunks defined in Appendix \ref{appendix:toy_retrieval}. The way to read this matrix is horizontally, row by row. For the \textit{neutral query} (first row), we find that it is closest to the \textit{neutral chunks}, followed by the \textit{male chunks}, and then, the \textit{female chunks}, as expected. We find the same ordering for the \textit{male query}. For the \textit{female query}, we can see that each of the \textit{female chunks} ranks the highest compared to their \textit{male} and \textit{neutral} counterparts, also as expected. Finally, as discovered and quantified in our experiments (Table \ref{table:method_2_results}), we can see that, due to the ``over-compensating effect", the \textit{debiasing query} is closest to the \textit{female chunks}, followed by \textit{neutral} and \textit{male}.

\begin{figure}[h!]
    \hspace{-0.7em}
    \begin{minipage}[b]{0.50\textwidth}
        \includegraphics[width=\textwidth]{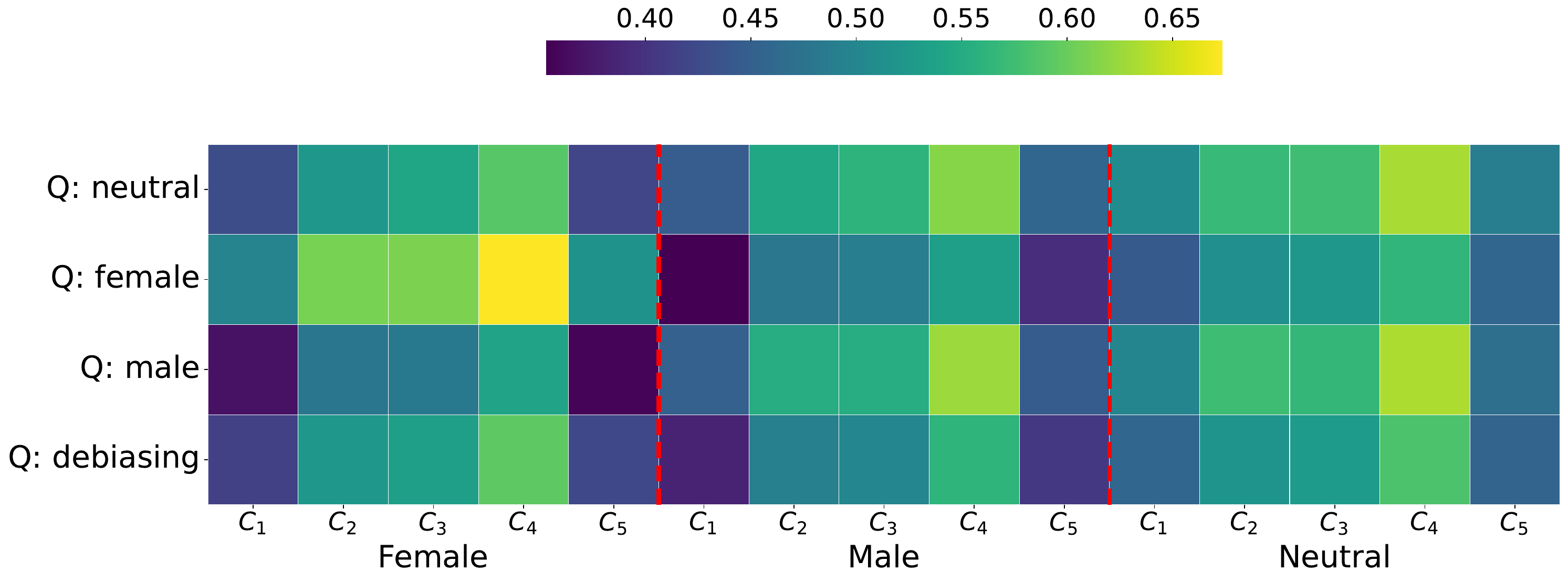}
    \end{minipage}
    \caption{:\hspace{0.1cm}Similarity matrix between the queries (y-axis) and the chunks (x-axis), using \textit{UAE-Large-V1} as embedding model.}
    \label{fig:similiraty_matrix}
\end{figure}

These results suggest that using a neutral query along with a top $k$ ranking, as is often the case in retrieval tasks, would miss out on the gendered chunks, whether male or female. Similarly when injecting debiasing context to the query, we would end-up with results closer to female. Instead of using top $k$, using an appropriate similarity threshold is difficult and more of an art than a science. To alleviate this issue, we propose a simple algorithm (see Algorithm 1), where we make use of our knowledge about the existence of bias in order to capture more of the relevant chunks.

\begin{algorithm}[h!]
\caption{Simple debiased retrieval algorithm for gender}
\label{alg:retrieval_algorithm}

\begin{algorithmic}[1]
\STATE \textbf{Input:} neutral query $q$, chunks $\{C_i | i=1,\ldots,n\}$, number of chunks to keep $k$, similarity metric $s(q, \cdot)$.
\STATE Transform \textit{neutral query} to \textit{female query}.
\STATE Retrieve top $k$ set: $\widetilde{S}=\{\widetilde{C}_i | i=1,\ldots,k\}$.
\STATE Transform \textit{neutral query} to \textit{male query}.
\STATE Retrieve top $m$ set: $\widehat{S}=\{\widehat{C}_i | i=1,\ldots,m\}$, where the lowest ranking match is $\widehat{C}_m = \argmin_{\widetilde{C} \in \widetilde{S}} s(q, \widetilde{C})$.

\STATE \textbf{Output:} $\widehat{S}$, $m$

\end{algorithmic}
\end{algorithm}

The algorithm first takes in a \textit{neutral} query. This query is converted to a \textit{positive} query and we retrieve the top $k$ chunks most similar to it. Then, we switch to a \textit{negative} query and retrieve all the chunks that have a higher similarity than the lowest ranked chunk present in the previous retrieval. This can be seen as dynamically choosing the final number of chunks that we retrieve, and will equally retrieve the \textit{neutral} and \textit{gendered} chunks. Note that this works because we assume that $s(\text{female query}, \text{related male chunk}) > s(\text{female query}, \text{unrelated chunk})$, which we show to be the case in Figure \ref{fig:similiraty_matrix_neutral_query} in Appendix \ref{appendix:retrieval}. This algorithm is readily applicable to any type of bias, but requires to be aware of the potential biases beforehand. Note that this is an initial concept and would require to be studied further in future work. In Appendix \ref{appendix:toy_retrieval} (Table \ref{tab:ranked_chunks}), we show the detailed results of using Algorithm \ref{alg:retrieval_algorithm} compared to simply using a regular neutral query, when applied to the gendered example depicted in Figure \ref{fig:similiraty_matrix}. In short, fixing $k$ to an arbitrary value such as 10 end-up missing out crucial relevant chunks. On the other hand, using our algorithm, we are able to capture all the 15 relevant chunks while leaving out random/non-related chunks.

\section{Conclusion}
\label{sec:conclusion}

Embedding models capture societal biases which are in turn perpetuated in downstream tasks and applications. It is thus important to study the extent and impact that biases have on those systems, and to understand how to mitigate the effects. In this paper, we discovered several new insights. Firstly, better performing model are more prone to capturing biases. Secondly, these models are largely better at following context by correctly capturing affirmative (non-neutral) semantics. However, surprisingly, we find that models are incapable of correctly embedding neutral semantics. Moreover, they tend to strongly embed neutral semantics as closest to the discriminated/biased against group. This behaviour corroborates the finding in the SugarCrepe++ dataset \cite{harsha2024sugarcrepe++} where the authors found that even SOTA models cannot correctly disentangle subtle semantics change to syntaxical overlap. Our work serves as another cautionary tale about the blind usage of such models in critical applications. Further, we have shown that the impact of gender bias on a toy retrieval task can be severe. In order to tackle issue, we introduced a simple algorithm which dynamically chooses the top $k$ number of results to return. In future work, we would like to investigate this algorithm further on more advanced scenarios. Finally, this analysis has been entirely performed on English text, and warrants investigating how multilingual models embed bias depending on the language. 

\section{Limitations}

This paper has a few limitations one should be aware of. These limitations were already mentioned in the paper but will be reiterated here. Firstly, quantifying biases requires us to be aware of them in the first place. This is not trivial as several biases may be unknown to us. Similarly, we may think that some of the beliefs that we hold are not biased, while other people may think otherwise. This issue is also relevant for debiasing embeddings: in our algorithm, we must be aware of which bias to take into consideration. Second, this study was strictly performed on English text. It is therefore limited to biases pertaining to perhaps more Western views. Generally, this is also the case when most of the training corpus comes from the internet whose content is Western dominated. It would thus be of interest to perform similar studies on non-English languages with multilingual embedders. Ultimately, as of today, there is no silver bullet method to remove the effects of biases from embeddings. Practitioners must be aware that biases are present and that applications will be affected.

\section{Ethical Impact}

We do not foresee any ethical impact from this study. Studies on biases will contain content that may be offensive to some readers. The authors of this paper do not agree with any kind of discrimination against any type of groups. In fact, we find it important to shed lights on the negative impact that biases in embeddings can have in AI-powered applications.


\clearpage

\bibliography{custom}

\appendix
\onecolumn
\section{Data}
\subsection{Concepts}
\label{appendix:data_geometry}

\vspace{0.5cm}

\begin{table}[h!]
\centering
\begin{tabular}{c|c|c}
\textbf{Gender} & \textbf{Age} & \textbf{Wealth} \\ \hline
she - he & old - young & poverty - wealth \\ 
daughter - son & elderly - youthful & destitute - affluent \\ 
her - him & aged - youthful & broke - prosperous \\ 
mother - father & senior - junior & needy - opulent \\ 
woman - man & ancient - modern & impoverished - luxurious \\ 
lady - gentleman & mature - immature & underprivileged - privileged \\ 
Jane Doe - John Doe & veteran - novice & indigent - well-off \\ 
girl - boy & old-timer - newcomer & bankrupt - flush \\ 
herself - himself & retiree - student & meager - abundant \\ 
female - male & granny - grandchild & bare - lavish \\ 
sister - brother & grandfather - grandson & humble - grand \\ 
wife - husband & grandmother - granddaughter & scant - ample \\ 
girlfriend - boyfriend & old man - young man & struggling - thriving \\ 
queen - king & old lady - young lady & frugal - extravagant \\ 
princess - prince & senior citizen - teenager & penniless - loaded \\ 
actress - actor & pensioner - child & modest - plush \\ 
niece - nephew & seniority - youth & cheap - expensive \\ 
aunt - uncle & old age - youth & sparse - abundant \\ 
bride - groom & elder - youngster & economical - spendthrift \\ 
mistress - master & elderliness - youthfulness & thrifty - opulent \\ 
 & grandparent - grandchild & skimping - splurging \\ 
 & centenarian - infant & frugality - exuberance \\ 
 &  & stingy - generous \\ 
 &  & lean - fat \\ 
 &  & subsistence - affluence \\ 
 &  & austere - sumptuous \\ 
 &  & threadbare - opulent \\ 
 &  & paltry - ample \\ 
 &  & deprived - wealthy \\ 
 &  & scraping - lavishing \\ 
 &  & measly - ample \\ 
 &  & scanty - plentiful \\ 
 &  & tight - flush \\ 
 &  & penurious - affluent \\ 
 &  & stripped - furnished \\ 
 &  & meagerly - luxuriously \\ 
 &  & barebones - elaborate \\ 
 &  & barren - fertile \\ 
 &  & deficient - sufficient \\ 
 &  & spare - plush \\ 
 &  & shortage - surplus \\ 
 &  & hand-to-mouth - prosperous \\ 
 &  & insufficient - ample \\ 
 &  & minimalist - luxurious \\ 
\end{tabular}
\caption{:\hspace{0.1cm}Chosen pairs for the concepts of \textit{gender}, \textit{age}, and \textit{wealth}.}
\label{tab:pairs_for_concepts}
\end{table}

\newpage

\subsection{Attributes and their Labels}
\vspace{0.1cm}

\begin{table}[ht!]
\centering
\fontsize{10}{11}\selectfont
\begin{tabular}{c|c|c}
\textbf{General occupations (gender)} & \textbf{Ethnicity (wealth)} & \textbf{Character traits (age)} 
\\ \hline
&  & \\
electrician: 0.89 & Bantu: 0.25 & wise: 0.8 \\ 
surgeon: 0.65 & Zulu: 0.3 & inflexible: 0.7 \\ 
police officer: 0.75 & Somali: 0.2 & knowledgeable: 0.6 \\ 
homemaker: 0.16 & Ashanti: 0.4 & tech-savvy: 0.1 \\ 
politician: 0.68 & Berber: 0.45 & forgetful: 0.6 \\ 
human resources: 0.28 & Tuareg: 0.45 & quick learner: 0.2 \\ 
miner: 0.94 & Uyghur: 0.3 & slow: 0.6 \\ 
beauty therapist: 0.13 & Tibetan: 0.55 & energetic: 0.2 \\ 
mechanic: 0.88 & Hmong: 0.4 & out of touch: 0.8 \\ 
writer: 0.56 & Indian: 0.7 & adaptable: 0.3 \\ 
construction worker: 0.84 & Latino: 0.6 & dependent: 0.5 \\ 
housekeeper: 0.15 & Caucasian: 0.8 & active: 0.2 \\ 
teacher: 0.39 & Greek: 0.75 & stuck in their ways: 0.7 \\ 
interior designer: 0.27 & Romanian: 0.6 & old-fashioned: 0.8 \\ 
captain: 0.79 & Scandinavian: 0.75 & vibrant: 0.2 \\ 
engineer: 0.68 & Persian: 0.7 & senile: 0.8 \\ 
bookkeeper: 0.44 & Kurdish: 0.6 & innovative: 0.3 \\ 
carpenter: 0.78 & Armenian: 0.65 & traditional: 0.7 \\ 
truck driver: 0.91 & Assyrian: 0.55 & dynamic: 0.3 \\ 
mathematician: 0.61 & Turkish: 0.6 & ambitious: 0.2 \\ 
librarian: 0.31 & Jewish: 0.8 & self-reliant: 0.5 \\ 
receptionist: 0.33 & Arab: 0.65 & forward-thinking: 0.3 \\ 
hairdresser: 0.27 & Mestizo: 0.45 & resourceful: 0.3 \\ 
CEO: 0.61 & Quechua: 0.4 & inept with technology: 0.8 \\ 
scientist: 0.53 & Mapuche: 0.3 & open-minded: 0.3 \\ 
firefighter: 0.73 & Mulatto: 0.45 & stubborn: 0.7 \\ 
pilot: 0.75 & Native American: 0.35 & rigid: 0.7 \\ 
philosopher: 0.60 & Cherokee: 0.3 & cynical: 0.8 \\ 
nurse: 0.23 & Navajo: 0.3 & settled: 0.7 \\ 
social worker: 0.40 & Sioux: 0.25 & \\ 
socialite: 0.46 & Inuit: 0.4 & \\ 
racer: 0.85 & Maori: 0.45 &  \\ 
administrative assistant: 0.34 & Aboriginal Australians: 0.35 &  \\ 
manager: 0.66 & Pacific islander: 0.55 &  \\ 
stylist: 0.27 &  &  \\ 
childcare provider: 0.21 &  &  \\ 
skipper: 0.88 &  &  \\ 
guidance counselor: 0.36 &  &  \\ 
accountant: 0.55 &  &  \\ 
warrior: 0.82 &  &  \\ 
computer scientist: 0.68 &  &  \\ 
broadcaster: 0.53 &  &  \\ 
gamer: 0.73 &  &  \\ 
plumber: 0.91 &  &  \\ 
architect: 0.53 &  &  \\ 
maestro: 0.81 &  &  \\ 
magician: 0.72 &  &  \\ 
nanny: 0.18 &  &  \\ 
 &  &  \\ 
\end{tabular}
\caption{:\hspace{0.1cm}Attributes and human-annotated labels. The labels correspond to the concepts specified in the column names. Labels were taken from 10 volunteers annotators and the mean was used as the final value. If needed, the labels were then rounded-off to two decimal place. The annotators were given the following instructions: give a continuous label, from 0 to 1, which reflects your personal opinion, where 0 is the closest to the discriminated against group (e.g., female for gender, poor for wealth, old for age) for each of the attributes in the three groups. The annotators were made aware of how these labels were going to be used in this study.}
\label{tab:bias_labels}
\end{table}

\newpage

\subsection{WEAT: Target and Attribute Sets}
\label{appendix:data_targets_attributes}

\vspace{0.2cm}

\textbf{WEAT 1:}

We use male and female target words with male-oriented occupations and female-oriented occupations as attributes. We have:

$T_1$ = \{she, daughter, her, mother, woman, lady, Jane Doe, girl, herself, female, sister, wife, girlfriend, queen, princess, actress, niece, aunt, bride, mistress\},

$T_2$ = \{he, son, him, father, man, gentleman, John Doe, boy, himself, male, brother, husband, boyfriend, king, prince, actor, nephew, uncle, groom, master\},

$A_1$ = \{homemaker, human resources, beauty therapist, housekeeper, teacher, interior designer, bookkeeper, librarian, receptionist, hairdresser, nurse, social worker, socialite, administrative assistant, stylist, childcare provider, guidance counselor, nanny\},

$A_2$ = \{electrician, surgeon, police officer, politician, miner, mechanic, writer, construction worker, captain, engineer, carpenter, truck driver, mathematician, ceo, scientist, firefighter, pilot, philosopher, racer, manager, skipper, accountant, warrior, computer scientist, broadcaster, gamer, plumber, architect, maestro, magician\}.

\vspace{0.2cm}

\textbf{WEAT 2:}

We use terms related to wealth (i.e., rich and poor) for our concept. For the attributes we use some major ethnic and cultural groups. We have:

$T_1$ = \{poverty, destitute, broke, needy, impoverished, underprivileged, indigent, bankrupt, meager, bare, humble, scant, struggling, frugal, penniless, modest, cheap, sparse, economical, thrifty, skimping, frugality, stingy, lean, subsistence, austere, threadbare, paltry, deprived, scraping, measly, scanty, tight, penurious, stripped, meagerly, barebones, barren, deficient, spare, shortage, hand-to-mouth, insufficient, minimalist\},

$T_2$ = \{wealth, affluent, prosperous, opulent, luxurious, privileged, well-off, flush, abundant, lavish, grand, ample, thriving, extravagant, loaded, plush, expensive, abundant, spendthrift, opulent, splurging, exuberance, generous, fat, affluence, sumptuous, opulent, ample, wealthy, lavishing, ample, plentiful, flush, affluent, furnished, luxuriously, elaborate, fertile, sufficient, plush, surplus, prosperous, ample, luxurious\},

$A_1$ = \{Bantu, Zulu, Somali, Ashanti, Berber, Tuareg, Uyghur, Hmong, Mestizo, Quechua, Mapuche, Mulatto, Native American, Cherokee, Navajo, Sioux, Inuit, Maori, Aboriginal Australians\}.

$A_2$ = \{ Tibetan, Indian, Latino, Caucasian, Greek, Romanian, Scandinavian, Persian, Kurdish, Armenian, Assyrian, Turkish, Jewish, Arab, Pacific islander\}.

\vspace{0.2cm}

\textbf{WEAT 3:}

We use terms related to old-age and young-age as target words. For attributes, we use character traits that are believe to be more related to older people and others more related to younger people. We have:

$T_1$ = \{old, elderly, aged, senior, ancient, mature, veteran, old-timer, retiree, granny, grandfather, grandmother, old man, old lady, senior citizen, pensioner, seniority, old age, elder, elderliness, grandparent, centenarian\},

$T_2$ = \{young, youthful, youthful, junior, modern, immature, novice, newcomer, student, grandchild, grandson, granddaughter, young man, young lady, teenager, child, youth, youngster, youthfulness, grandchild, infant\},

$A_1$ = \{wise, inflexible, knowledgeable, forgetful, slow, out-of-touch, dependent, stuck in their ways, old-fashioned, senile, traditional, self-reliant, inept with technology, stubborn, rigid, cynical, settled\},

$A_2$ = \{tech-savvy, quick learner, energetic, adaptable, active, vibrant, innovative, dynamic, ambitious, forward-thinking, resourceful, open-minded, adaptive, innocent\}.

\vspace{0.2cm}

\newpage

\subsection{Prompt templates}
\label{appendix:prompts}

\vspace{1cm}

\begin{table}[h!]
\centering
\begin{tabular}{cc} 
    \begin{minipage}{0.47\textwidth} 
        \centering
        \hspace{-1.9em}
        \begin{tabular}{p{1.5cm}p{5.8cm}}
        \specialrule{1.3pt}{0pt}{0pt} 
        \textbf{Scenarios} & \multicolumn{1}{c}{\textbf{Text to embed}} \\
        \specialrule{0.5pt}{0pt}{0pt} 
        \multirow{3}{=}{\textbf{Neutral}} & I want to find information about a person. This person's character trait is \{attribute\}. \\
        \specialrule{0.5pt}{0pt}{0pt} 
        \multirow{4}{=}{\textbf{Debiasing}} & I want to find information about a person. This person's character trait is \{attribute\}. This person's age is unknown. \\
        \specialrule{0.5pt}{0pt}{0pt} 
        \multirow{3}{=}{\textbf{Positive}} & I want to find information about a person. This person's character trait is \{attribute\}. This person is old. \\
        \specialrule{0.5pt}{0pt}{0pt} 
        \multirow{3}{=}{\textbf{Negative}} & I want to find information about a person. This person's character trait is \{attribute\}. This person is young. \\
        \specialrule{1.3pt}{0pt}{0pt} 
        \end{tabular}
        \caption{:\hspace{0.1cm}Example of the four scenarios, for the \textit{age} concept. Note that the use of ``positive'' and ``negative'' is arbitrary and not to be taken literally.}
        \label{table:scenarios_age}
    \end{minipage}
    &
    \begin{minipage}{0.47\textwidth} 
        \centering
        \begin{tabular}{p{1.5cm}p{5.8cm}}
        \specialrule{1.3pt}{0pt}{0pt} 
        \textbf{Scenarios} & \multicolumn{1}{c}{\textbf{Text to embed}} \\
        \specialrule{0.5pt}{0pt}{0pt} 
        \multirow{3}{=}{\textbf{Neutral}} & I want to find information about a person. This person is of \{attribute\} ethnicity. \\
        \specialrule{0.5pt}{0pt}{0pt} 
        \multirow{4}{=}{\textbf{Debiasing}} & I want to find information about a person. This person is of \{attribute\} ethnicity. This person's social status is unknown. \\
        \specialrule{0.5pt}{0pt}{0pt} 
        \multirow{3}{=}{\textbf{Positive}} & I want to find information about a person. This person is of \{attribute\} ethnicity. This person is poor. \\
        \specialrule{0.5pt}{0pt}{0pt} 
        \multirow{3}{=}{\textbf{Negative}} & I want to find information about a person. This person is of \{attribute\} ethnicity. This person is wealthy. \\
        \specialrule{1.3pt}{0pt}{0pt} 
        \end{tabular}
        \caption{:\hspace{0.1cm}Example of the four scenarios, for the \textit{wealth} concept. Note that the use of ``positive'' and ``negative'' is arbitrary and not to be taken literally.}
        \label{table:scenarios_wealth}
    \end{minipage}
\end{tabular}
\end{table}

\section{Area Under the Curve}
\label{appendix:auc}

\begin{figure}[ht!]
    \centering
    \begin{minipage}[b]{0.46\textwidth}
        \centering
        \includegraphics[width=\textwidth]{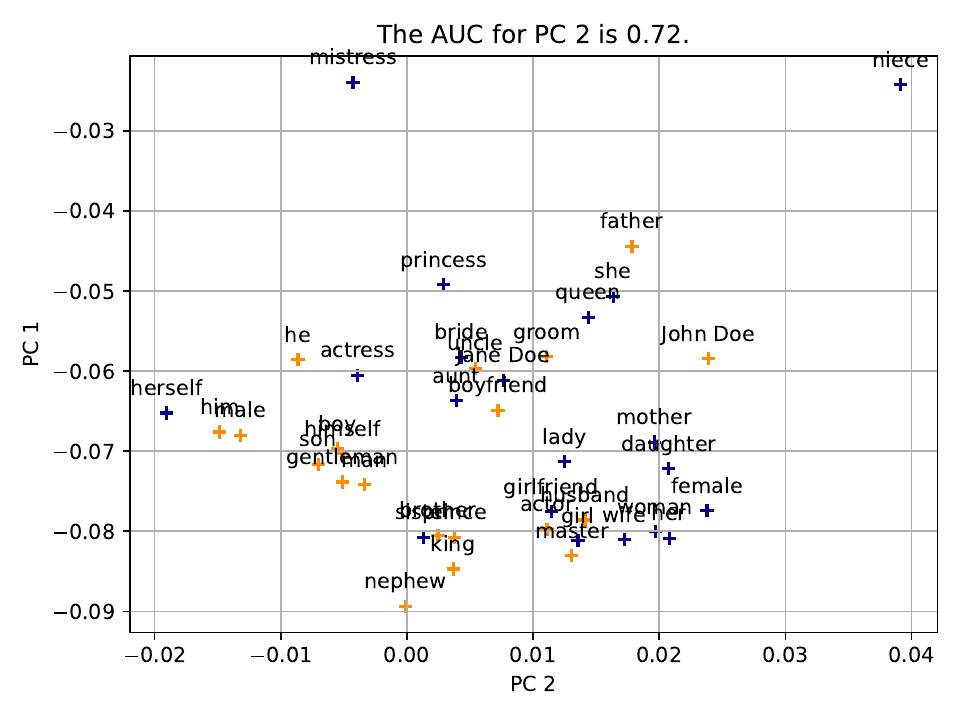}
        \makebox[\textwidth][c]{\hspace{1cm}\small(a) FacebookAI-xlm-roberta-base.}
    \end{minipage}
    \hfill
    \begin{minipage}[b]{0.46\textwidth}
        \centering
        \includegraphics[width=\textwidth]{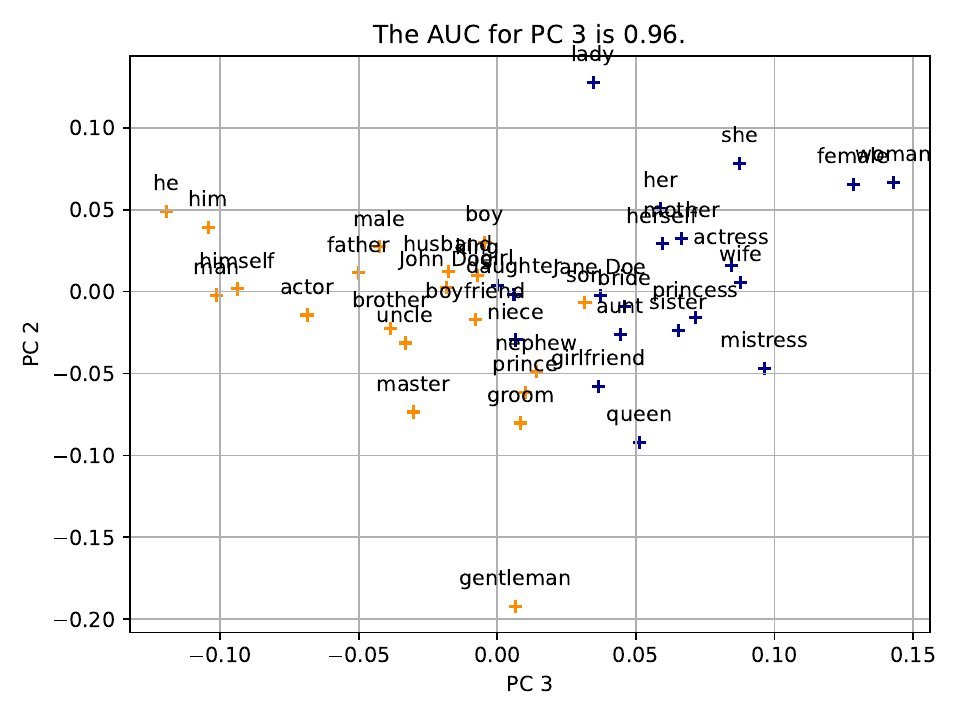}
        \makebox[\textwidth][c]{\hspace{1cm}\small(b) Intfloat-multilingual-e5-large-instruct.}  
    \end{minipage}

    \caption{:\hspace{0.1cm}Area under the curve achieved for the principal component with the largest AUC, displayed as the x-axis, for the gender concept. The component on the y-axis is used only for plotting purposes and not used in the AUC computation. On the left, the AUC is low, resulting in a poor linear separation of the two classes (yellow and blue). On the other hand, on the right, PC 3 is able to linearly separate the gender terms.}
    \label{fig:auc_gender}
\end{figure}

\begin{figure}[h!]
    \centering
    \begin{minipage}[b]{0.46\textwidth}
        \centering
        \includegraphics[width=\textwidth]{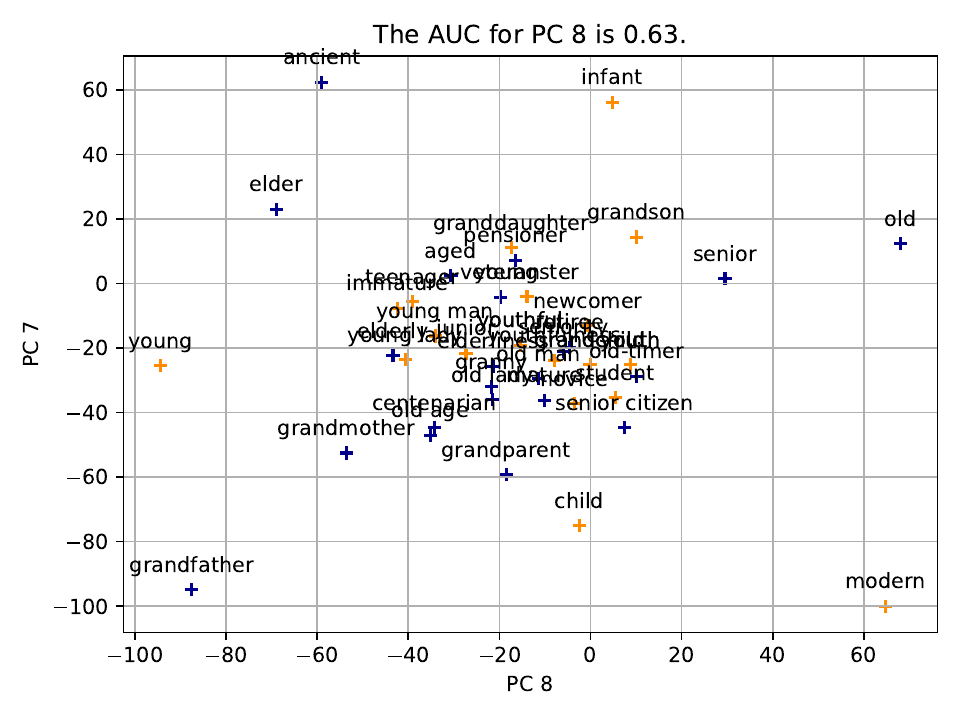}
        \makebox[\textwidth][c]{\hspace{1cm}\small(a) Mistral.}  
    \end{minipage}
    \hfill
    \begin{minipage}[b]{0.46\textwidth}
        \centering
        \includegraphics[width=\textwidth]{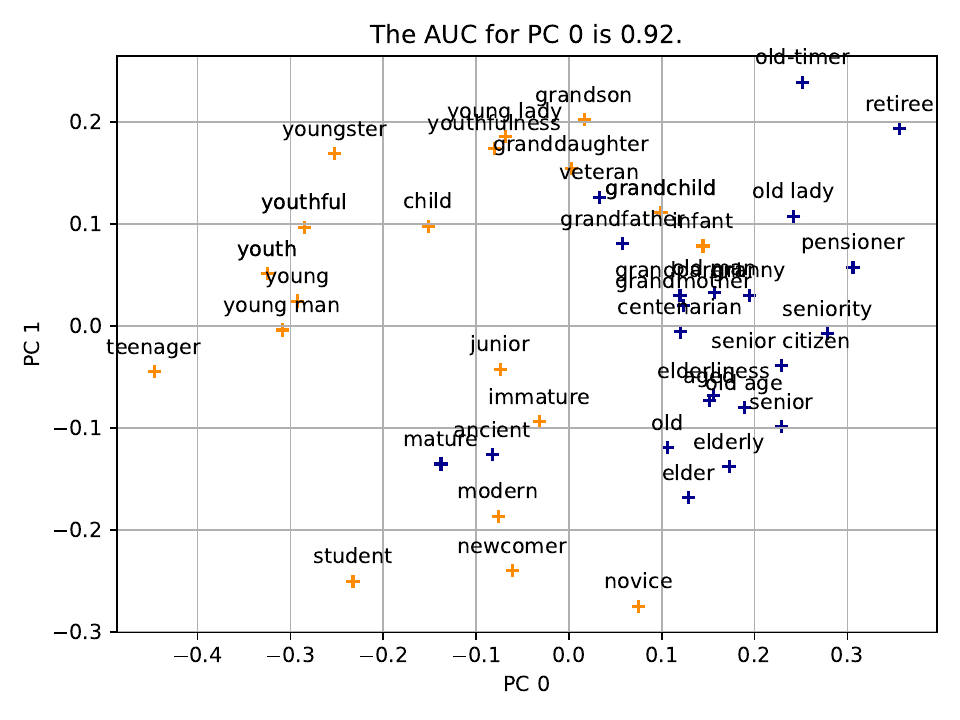}
        \makebox[\textwidth][c]{\hspace{1cm}\small(b) Alibaba-NLP-gte-large-en-v1.5.}  
    \end{minipage}

    \caption{:\hspace{0.1cm}Area under the curve achieved for the principal component with the largest AUC, displayed as the x-axis, for the age concept. The component on the y-axis is used only for plotting purposes and not used in the AUC computation. On the left, the AUC is low, resulting in a poor linear separation of the two classes (yellow and blue). On the other hand, on the right, PC 0 is able to separate the age terms much better.}
    \label{fig:auc_age}
\end{figure}

In this section, we display some examples of learned concepts for various embedding models. We show cases where the best direction is not given by the first eigenvector $b_1$, as was assumed in \cite{bolukbasi2016man}. We also exhibit cases where the concept is not well represented by the resulting concept direction $g$ (i.e., AUC $<$ 0.8). In Figure \ref{fig:auc_gender} (gender concept) and Figure \ref{fig:auc_age} (age concept), we display the learned concept direction $g$ on the x-axis, for the model with the lowest AUC (on the left) and model with the highest AUC (on the right).

\newpage

\section{Correlation to Human-Annotated Labels}
\label{appendix:correlation_labels}
\vspace{-0.3cm}
\begin{figure}[h!]
    \centering
    \begin{minipage}[b]{0.45\textwidth}
        \centering
        \includegraphics[width=\textwidth]{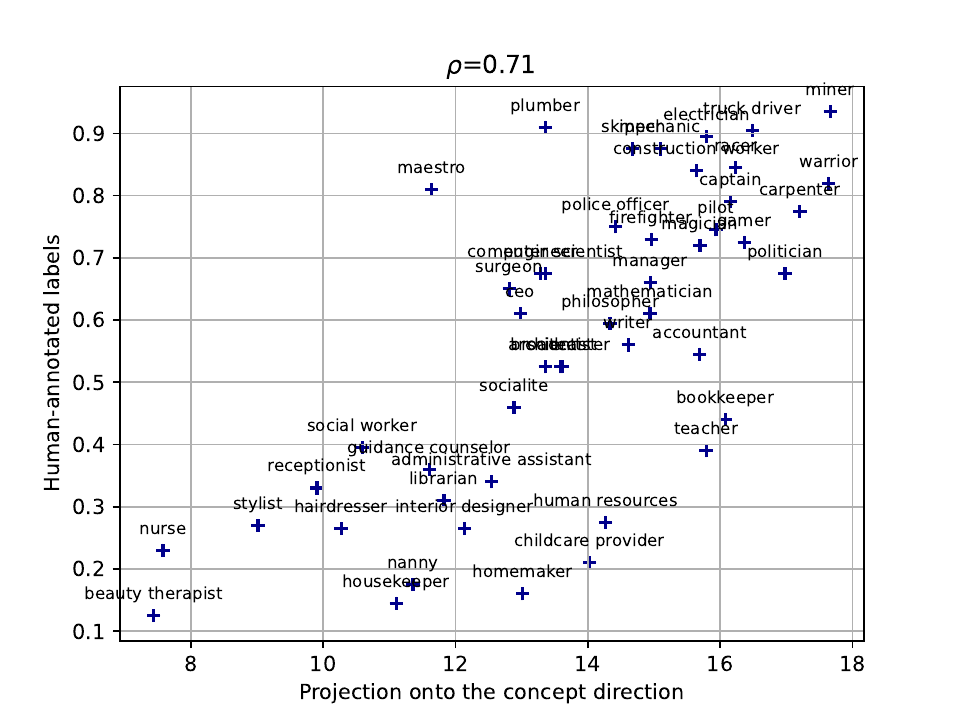}
        \makebox[\textwidth][c]{\parbox{\textwidth}{\small(a) Correlation between human-labeled bias and concept projection, for the \textit{neutral} context.}}
    \end{minipage}
    \hfill
    \begin{minipage}[b]{0.45\textwidth}
        \centering
        \includegraphics[width=\textwidth]{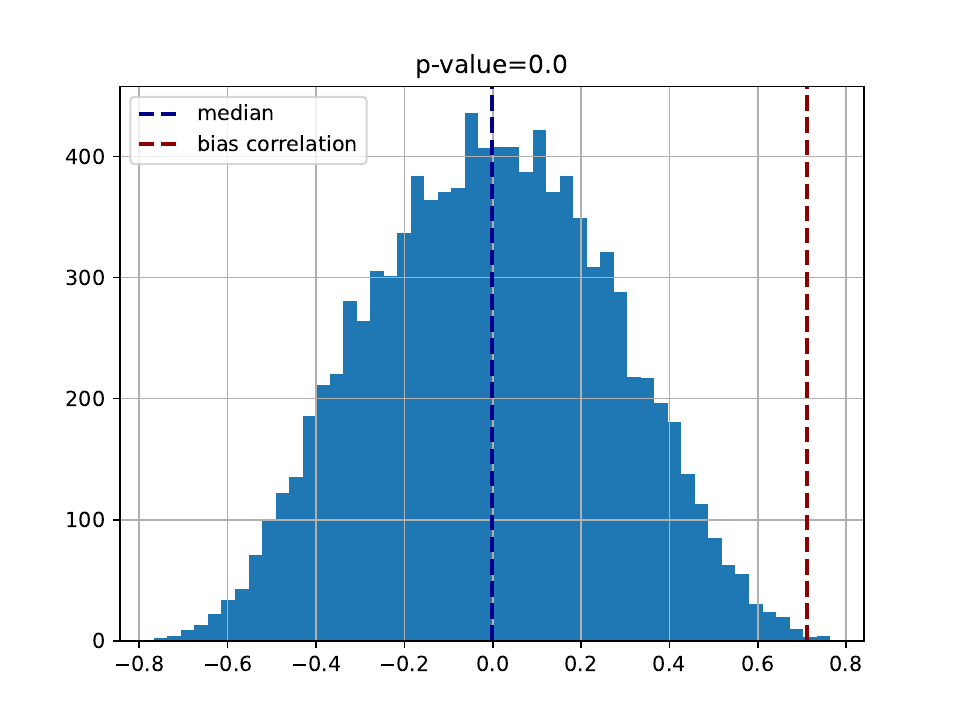}
        \makebox[\textwidth][c]{\parbox{\textwidth}{\small(b) Correlations for random projections (blue) and projection onto the gender concept, for the \textit{neutral} context.}}
    \end{minipage}

    \vskip\baselineskip
    \begin{minipage}[b]{0.45\textwidth}
        \centering
        \includegraphics[width=\textwidth]{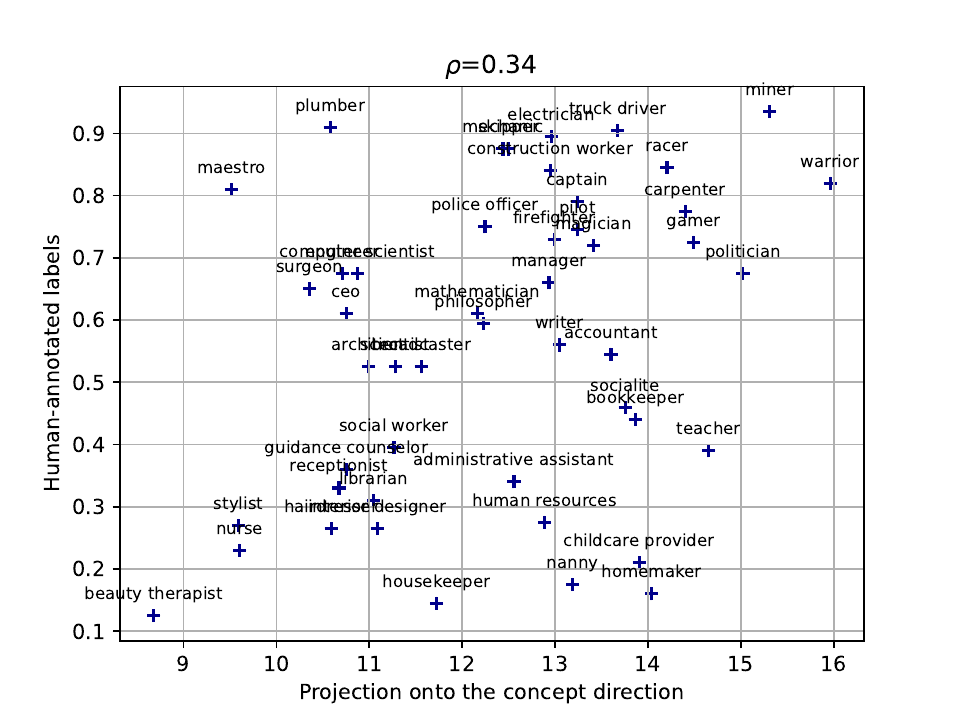}
        \makebox[\textwidth][c]{\parbox{\textwidth}{\small(c) Correlation between human-labeled bias and concept projection, for the \textit{debiasing} context.}}
    \end{minipage}
    \hfill
    \begin{minipage}[b]{0.45\textwidth}
        \centering
        \includegraphics[width=\textwidth]{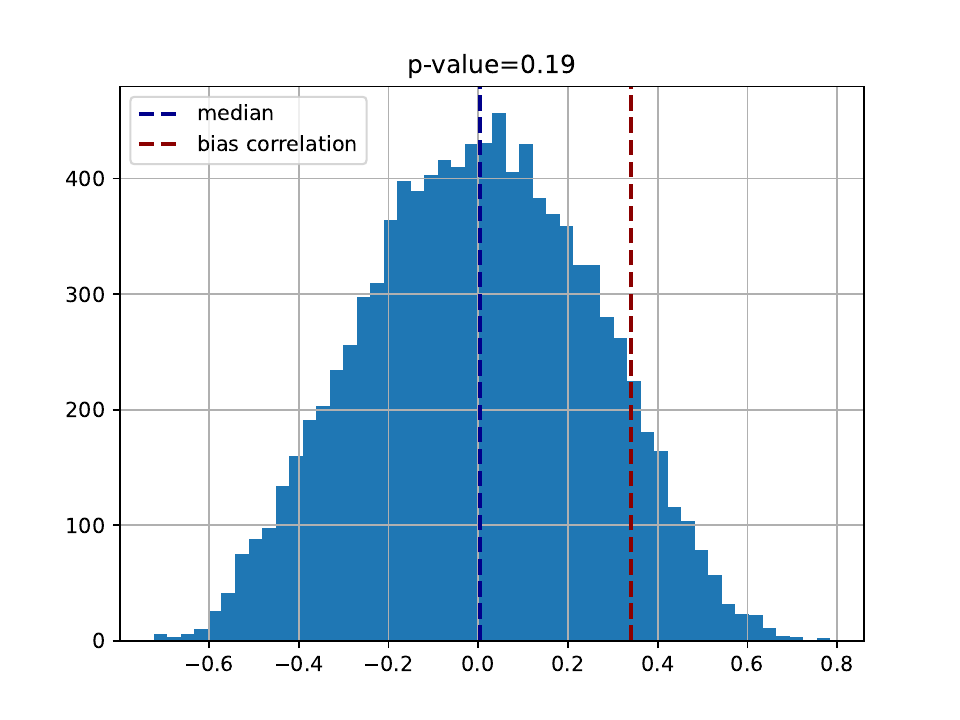}
        \makebox[\textwidth][c]{\parbox{\textwidth}{\small(d) Correlations for random projections (blue) and projection onto the gender concept, for the \textit{debiasing} context.}}
    \end{minipage}
\end{figure}
\newpage
\begin{figure}[htb!]
    \centering
    \begin{minipage}[b]{0.46\textwidth}
        \centering
        \includegraphics[width=\textwidth]{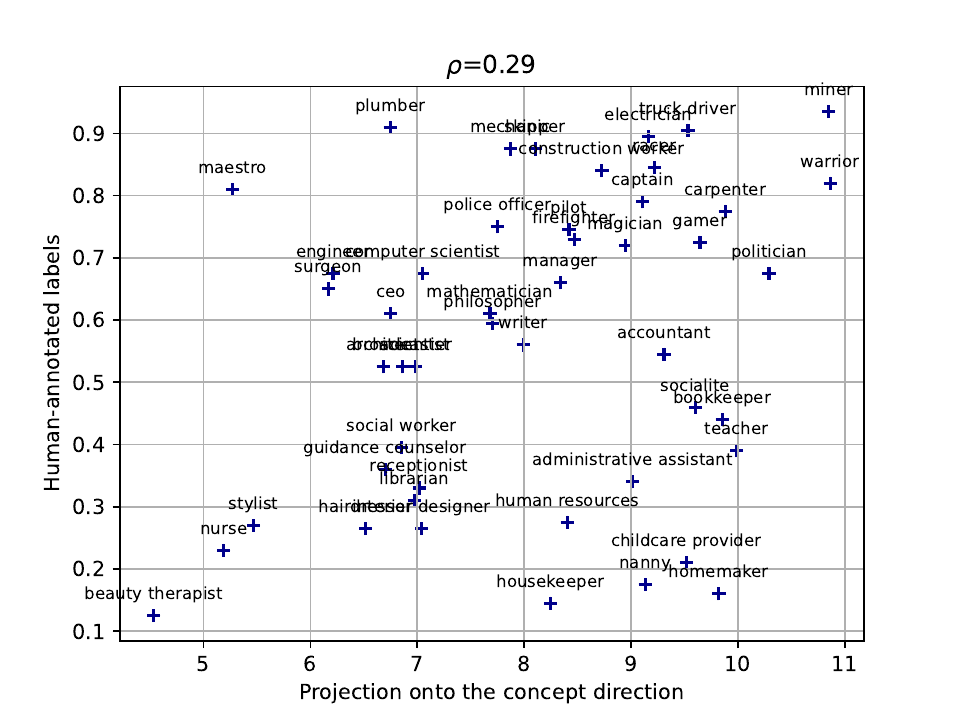}
        \makebox[\textwidth][c]{\parbox{\textwidth}{\small(e) Correlation between human-labeled bias and concept projection, for the \textit{positive} context.}}
    \end{minipage}
    \hfill
    \begin{minipage}[b]{0.46\textwidth}
        \centering
        \includegraphics[width=\textwidth]{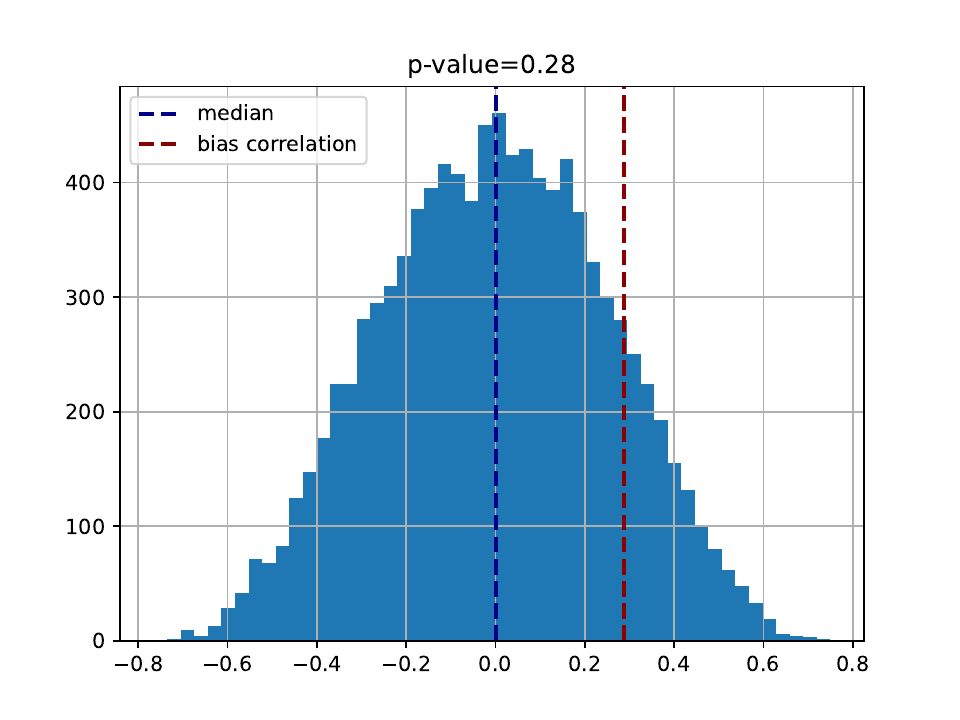}
        \makebox[\textwidth][c]{\parbox{\textwidth}{\small(f) Correlations for random projections (blue) and projection onto the gender concept (red), for the \textit{positive} context.}}
    \end{minipage}

    \caption{:\hspace{0.1cm}Results for the \textit{gender} concept and the \textit{occupations} attribute for the \textit{positive} context. The correlation obtained using the projection onto the gender direction is compared to the correlation obtained with $10^4$ random projections to compute a p-value. Note that for Figure (e), the values on the x-axis are more shifted towards the female gender due to the effect of context.}
    \label{fig:geometry_correlations2}
\end{figure}

\section{Detailed Results for the Geometry-based Method}
\label{sec:appendix_method1}

\begin{table*}[h!]
\centering
\resizebox{0.9\textwidth}{!}{
\begin{tabular}{lllllllllll}
\toprule
 \multicolumn{3}{c}{\shortstack{\textbf{Concept = gender}, \\ \textbf{Attributes = occupations}}} & \multicolumn{3}{c}{\shortstack{\textbf{Concept = wealth}, \\ \textbf{Attributes = ethnicities}}} &  \multicolumn{3}{c}{\shortstack{\textbf{Concept = age}, \\ \textbf{Attributes = character traits}}}\\
\cmidrule(lr){1-3} \cmidrule(lr){4-6} \cmidrule(lr){7-9} 
 \textbf{neutral} & \textbf{debiasing} & \textbf{positive} & \textbf{neutral} & \textbf{debiasing} & \textbf{positive} &  \textbf{neutral} & \textbf{debiasing} & \textbf{positive} \\
\midrule

$0.66^{***}$ & $0.52^{**}$ & $0.20$ & $0.53^{**}$ & $0.49^{*}$ & $0.14$ &$0.72^{***}$& $0.75^{***}$&$0.64^{**}$\\
$0.83^{***}$ & $0.73^{***}$ & $0.41$ & $0.40$ & $0.15$ & $0.11$ & $0.55^{**}$&$0.55^{**}$&$0.05$\\
$0.66^{***}$ & $0.12$ & $0.18$ & $0.56^{**}$ & $0.51^{*}$ & $0.27$ &$0.60^{**}$&$0.43$&$0.34$\\
$0.61^{***}$ & $0.02$ & $0.09$ & $0.49^{**}$ & $0.38$ & $0.17$ &$0.58^{**}$&$0.40$&$0.30$\\
$0.74^{***}$ & $0.64^{***}$ & $0.40$ & $0.55^{**}$ & $0.64^{***}$ & $0.53^{**}$ &$0.71^{***}$&$0.76^{***}$&$0.59^{**}$\\
$0.71^{***}$ & $0.34$ & $0.28$ & $0.18$ & $0.05$ & $0.18$ &$0.51^{**}$&$0.37$&$0.19$\\
$0.54^{**}$ & $0.22$ & $0.21$ & $0.52^{**}$ & $0.47^{*}$ & $0.29$ &$0.55^{**}$&$0.39$&$0.31$ \\
$0.54^{**}$ & $0.35$ & $0.37$ & $0.52^{**}$ & $0.45^{*}$ & $0.29$ &$0.54^{**}$&$0.37$&$0.29$\\
$0.17$ & $0.33$ & $0.27$ & $0.26$ & $0.30$ & $0.31$ &$0.63^{***}$&$0.55$&$0.54$\\
$0.44$ & $0.34$ & $0.26$ & $0.17$ & $0.06$ & $0.05$ &$0.24$&$0.03$&$0.02$\\
$0.79^{***}$ & $0.54^{**}$ & $0.36$ & $0.25$ & $0.05$ & $0.09$ & $0.77^{***}$&$0.74^{***}$&$0.63^{**}$\\
$0.01$ & $0.12$ & $0.09$ & $0.09$ & $0.04$ & $0.09$ &$0.02$&$0.05$&$0.05$\\
$0.42$ & $0.32$ & $0.29$ & $0.03$ & $0.21$ & $0.20$ &$0.19$&$0.12$&$0.11$\\
$0.31$ & $0.46^{**}$ & $0.38$ & $0.04$ & $0.42^{*}$ & $0.49^{**}$ &$0.17$&$0.20$&$0.13$\\

$0.40$ & $0.32$ & $0.38$ & $0.31$ & $0.32$ & $0.19$ &$0.07$&$0.06$&$0.04$\\
$0.81^{***}$ & $0.76^{***}$ & $0.43$ & $0.45^{*}$ & $0.29$ & $0.02$ &$0.15$&$0.05$&$0.09$\\
$0.0$ & $0.0$ & $0.52^{**}$ & $0.62^{**}$ & $0.02$ & $0.17$ &$0.14$&$0.30$&$0.36$\\
$0.17$ & $0.25$ & $0.29$ & $0.66^{***}$ & $0.15$ & $0.16$ &$0.36$&$0.22$&$0.46$\\
$0.58^{***}$ & $0.45^{**}$ & $0.61$ & $0.13$ & $0.64^{**}$ & $0.38$ &$0.03$&$0.10$&$0.23$\\
\bottomrule
\end{tabular}
}
\caption{:\hspace{0.1cm}Absolute correlation $|\rho|$, as described in Equation \ref{eq:bias_corr}, computed for three different contexts (neutral, debiasing, positive), across 3 pairs of concepts and attributes. The associated p-value (see Equation \ref{eq:p_value_1}) is given by the asterisk notation. For a confidence level of $1 - \alpha$, statistical significance is denoted as follows: $^{*}, ^{**}, ^{***}$, for $\alpha$ = $10\%$, $5\%$, and $1\%$, respectively.}
\label{table:method_1_results}
\end{table*}

\newpage

\section{Queries and Chunks for Toy Retrieval Task}
\label{appendix:toy_retrieval}
\subsection{Chunks}
\vspace{0.5cm}
\begin{table}[htb!]
    \hspace{-1.4em}
    \begin{tabular}{p{1.45cm}|p{14.5cm}}
        \hline
        \specialrule{1.3pt}{0pt}{0pt} 
        \textbf{Scenarios} & \multicolumn{1}{c}{\textbf{Text to embed}} \\
        \specialrule{0.5pt}{0pt}{0pt} 
        \multirow{7}{=}{\textbf{Female}} 
        & $C_1$: She commanded the largest naval fleet in the history of the nation. \\
        & $C_2$: The colonel devised a successful counterinsurgency strategy. She turned the tide of the conflict. \\ 
        & $C_3$: The officer was promoted to brigadier general after her decisive victory in the northern territories. \\
        & $C_4$: The general authored a highly regarded manual on modern warfare tactics. A brilliant lady! \\ 
        & $C_5$: She oversaw the construction of the country’s most advanced aircraft carrier. \\ \hline
        \multirow{7}{=}{\textbf{Male}} 
        & $C_1$: He commanded the largest naval fleet in the history of the nation. \\ 
        & $C_2$: The colonel devised a successful counterinsurgency strategy. He turned the tide of the conflict. \\
        & $C_3$: The officer was promoted to brigadier general after his decisive victory in the northern territories. \\ 
        & $C_4$: The general authored a highly regarded manual on modern warfare tactics. A brilliant sir! \\ 
        & $C_5$: He oversaw the construction of the country’s most advanced aircraft carrier. \\ \hline
        \multirow{7}{=}{\textbf{Neutral}} 
        & $C_1$: This person commanded the largest naval fleet in the history of the nation. \\ 
        & $C_2$: The colonel devised a successful counterinsurgency strategy. This person turned the tide of the conflict. \\ 
        & $C_3$: The officer was promoted to brigadier general after this person's decisive victory in the northern territories. \\
        & $C_4$: The general authored a highly regarded manual on modern warfare tactics. A brilliant person! \\ 
        & $C_5$: This person oversaw the construction of the country’s most advanced aircraft carrier. \\ \hline
        \multirow{7}{=}{\textbf{Random}} 
        & $C_1$: A cat stretched lazily on the windowsill, basking in the warmth of the afternoon sun. \\
        & $C_2$: The train rattled along the tracks, carrying passengers through the misty countryside. \\
        & $C_3$: A musician played his guitar under the streetlight, his melodies echoing through the quiet night. \\
        & $C_4$: The chef chopped vegetables with precision, the sound of the knife rhythmic against the cutting board. \\
        & $C_5$: A young couple walked hand in hand along the beach, the waves gently lapping at their feet. \\ \hline
    \end{tabular}
    \caption{:\hspace{0.1cm}Simple sentences used in a retrieval task. For \textit{Female}, \textit{Male}, and \textit{Neutral}, the sentences carry the same meaning and only differ in their gender information. The remaining 5 sentences under \textit{Random} are unrelated sentences to the query.}
    \label{tab:sentences}
\end{table}

\subsection{Queries}
\vspace{0.1cm}
\begin{table}[h!]
    \centering
    \begin{tabular}{p{1.5cm}|p{14cm}}
        \hline
        \specialrule{1.3pt}{0pt}{0pt} 
        \textbf{Scenarios} & \multicolumn{1}{c}{\textbf{Text to embed}} \\
        \specialrule{0.5pt}{0pt}{0pt} 
        \multirow{2}{=}{\textbf{Female}} 
        & I want to find information about a high-ranking personnel in the army. 
          This person is a female. \\
        \multirow{1}{=}{\textbf{Male}} 
        & I want to find information about a high-ranking personnel in the army. 
          This person is a male. \\
        \multirow{1}{=}{\textbf{Neutral}} 
        & I want to find information about a high-ranking personnel in the army. \\
        \multirow{2}{=}{\textbf{Debiasing}} 
        & I want to find information about a high-ranking personnel in the army.
        This person's gender is not known. \\ \hline
    \end{tabular}
    \caption{:\hspace{0.1cm}Simple queries used in a retrieval task. The sentences carry the same meaning and only differ in their gender information.}
    \label{tab:queries}
\end{table}

\subsection{Retrieval Results}
\label{appendix:retrieval}

In this section, we display the results of a simple retrieval scenario, described in Section \ref{sec:toy_rag}, using the \textit{WhereIsAI/UAE-Large-V1} embedding model.

\begin{figure}[h!]
    \centering
    \begin{minipage}[b]{0.89\textwidth}
        \centering
        \includegraphics[width=\textwidth]{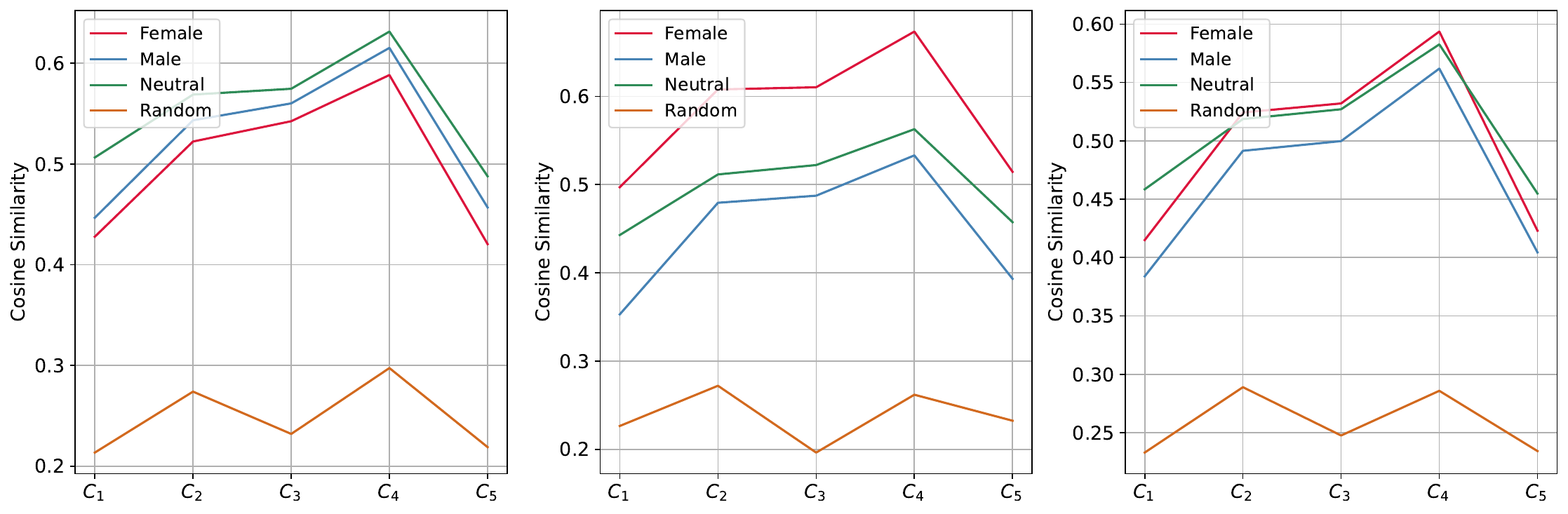} 
    \end{minipage}

    \caption{:\hspace{0.1cm}Cosine similarity between the queries defined in Table \ref{tab:queries} and the chunks in Table \ref{tab:sentences}. \textbf{Left:} \textit{neutral query}. \textbf{Center:} \textit{female query}. \textbf{Right:} \textit{debiasing query}}
    \label{fig:similiraty_matrix_neutral_query}
\end{figure}

\begin{table}[h!]
    \centering
    \begin{tabular}{c|c|c}
        \hline
        \textbf{Neutral} & \textbf{Female} & \textbf{Male} \\
        \hline
        neutral $C_4$ & female $C_4$ & neutral $C_4$ \\
        male $C_4$ & female $C_3$ & male $C_4$ \\
        female $C_4$ & female $C_2$ & neutral $C_2$ \\
        neutral $C_3$ & neutral $C_4$ & neutral $C_3$ \\
        neutral $C_2$ & male $C_4$ & male $C_3$ \\
        male $C_3$ & neutral $C_3$ & male $C_2$ \\
        male $C_2$ & female $C_5$ & female $C_4$ \\
        female $C_3$ & neutral $C_2$ & neutral $C_1$ \\
        female $C_2$ & female $C_1$ & female $C_3$ \\
        neutral $C_1$ & male $C_3$ & female $C_2$ \\
        neutral $C_5$ & male $C_2$ & neutral $C_5$ \\
        male $C_5$ & neutral $C_5$ & male $C_1$ \\
        male $C_1$ & neutral $C_1$ & male $C_5$ \\
        female $C_1$ & male $C_5$ & female $C_1$ \\
        female $C_5$ & male $C_1$ & female $C_5$ \\
        random $C_4$ & random $C_2$ & random $C_4$ \\
        random $C_2$ & random $C_4$ & random $C_2$ \\
        random $C_3$ & random $C_5$ & random $C_3$ \\
        random $C_5$ & random $C_1$ & random $C_1$ \\
        random $C_1$ & random $C_3$ & random $C_5$ \\
        \hline
    \end{tabular}
    \caption{:\hspace{0.1cm}Ranking of each chunk according their cosine similarity with the corresponding query (shown in each column).}
    \label{tab:ranked_chunks}
\end{table}

We see that for a simple top 10 retrieval strategy using the \textit{neutral query}, we would miss out on the 5 folowing relevant chunks: \textit{neutral $C_5$}, \textit{male $C_5$}, \textit{male $C_1$}, \textit{female $C_1$}, and \textit{female $C_1$}. Instead, using our algorithm (Algorithm \ref{alg:retrieval_algorithm}), as a first step we retrieve the top 10 results from the second column (from \textit{female $C_4$} to \textit{male $C_3$}). Then, we retrieve all the results from the third column, that are ranked higher than the lowest chunk (included) retrieved in the previous step. In this case, the lowest chunk is \textit{female $C_5$}. Therefore, the final chunks retrieved when using the algorithm correspond to all the relevant chunks without any of the random ones.

\end{document}